\documentclass[10pt,twocolumn,letterpaper]{article}

\usepackage{cvpr}
\usepackage{times}
\usepackage{epsfig}
\usepackage{graphicx}
\usepackage{amsmath}
\usepackage{amssymb}
\usepackage{booktabs}
\usepackage{subfig}
\usepackage{tabularx}
\usepackage{eso-pic}
\usepackage{xspace}
\usepackage{footnote}
\usepackage{mathtools}
\usepackage[super]{nth}
\usepackage{amssymb}
\usepackage{pifont}
\usepackage[dvipsnames]{xcolor}
\usepackage[pagebackref=true,breaklinks=true,letterpaper=true,colorlinks,bookmarks=false]{hyperref}

\usepackage[breaklinks=true,bookmarks=false]{hyperref}

\cvprfinalcopy 

\def\cvprPaperID{710} 

\begin{document}

\title{AdaCoF: Adaptive Collaboration of Flows for Video Frame Interpolation}

\author{Hyeongmin Lee$^{1}$ \quad
	Taeoh Kim$^{1}$ \quad
	Tae-young Chung$^{1}$ \quad
	Daehyun Pak$^{1}$ \quad
	Yuseok Ban$^{2}$ \quad
	Sangyoun Lee$^{1*}$ \\ \vspace{0.01cm}
	\and 
	$^{1}$Yonsei University \\
	{\tt\small \{minimonia,kto,tato0220,koasing,syleee\}@yonsei.ac.kr}
	\and
	$^{2}$Agency for Defense Development \\
	{\tt\small ban@add.re.kr}
}

\maketitle

\begin{abstract}
	Video frame interpolation is one of the most challenging tasks in video processing research. Recently, many studies based on deep learning have been suggested. Most of these methods focus on finding locations with useful information to estimate each output pixel using their own frame warping operations. However, many of them have Degrees of Freedom~(DoF) limitations and fail to deal with the complex motions found in real world videos. To solve this problem, we propose a new warping module named Adaptive Collaboration of Flows~(AdaCoF). Our method estimates both kernel weights and offset vectors for each target pixel to synthesize the output frame. AdaCoF is one of the most generalized warping modules compared to other approaches, and covers most of them as special cases of it. Therefore, it can deal with a significantly wide domain of complex motions. To further improve our framework and synthesize more realistic outputs, we introduce dual-frame adversarial loss which is applicable only to video frame interpolation tasks. The experimental results show that our method outperforms the state-of-the-art methods for both fixed training set environments and the Middlebury benchmark. Our source code is available at \url{https://github.com/HyeongminLEE/AdaCoF-pytorch}.
\end{abstract}

\section{Introduction}

Synthesizing the intermediate frame when consecutive frames have been provided is one of the main research topics in the video processing area. Using a frame interpolation algorithm, we can obtain slow-motion videos from ordinary videos without using professional high-speed cameras. In addition, we can freely convert the frame rates of the videos so it can be applied to the video coding system. To interpolate the intermediate frame of a video requires an understanding of motion, unlike image pixel interpolation. Unfortunately, real world videos contain not only simple motions, but also large and complex ones, making the task significantly more difficult.

\begin{figure}
	\setlength{\belowcaptionskip}{-24pt}
	\begin{center}
		\subfloat[The Kernel-Based Approach]
		{\includegraphics[width=0.46\linewidth]{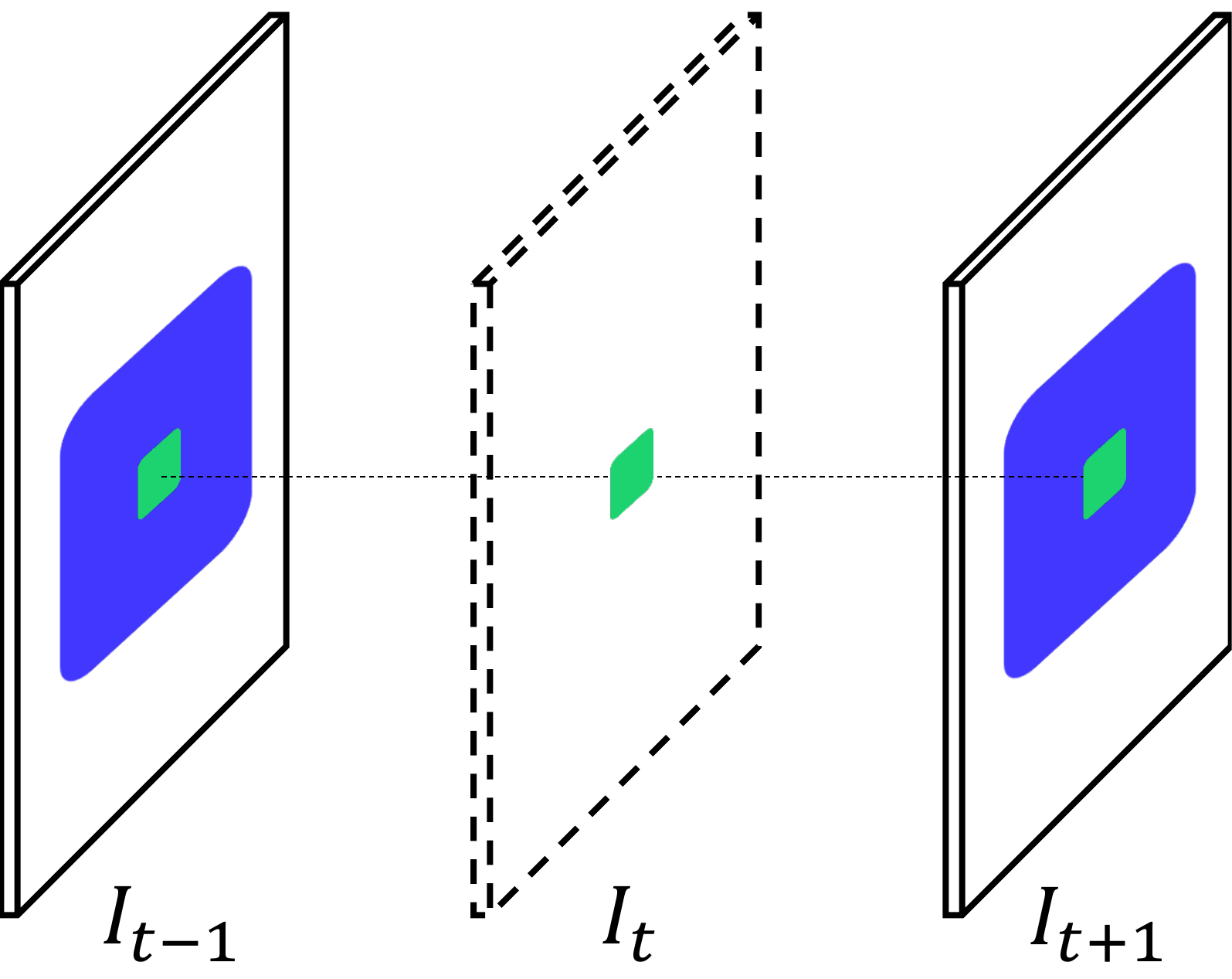}}\,
		\hfill
		\subfloat[The Flow-Based Approach]
		{\includegraphics[width=0.46\linewidth]{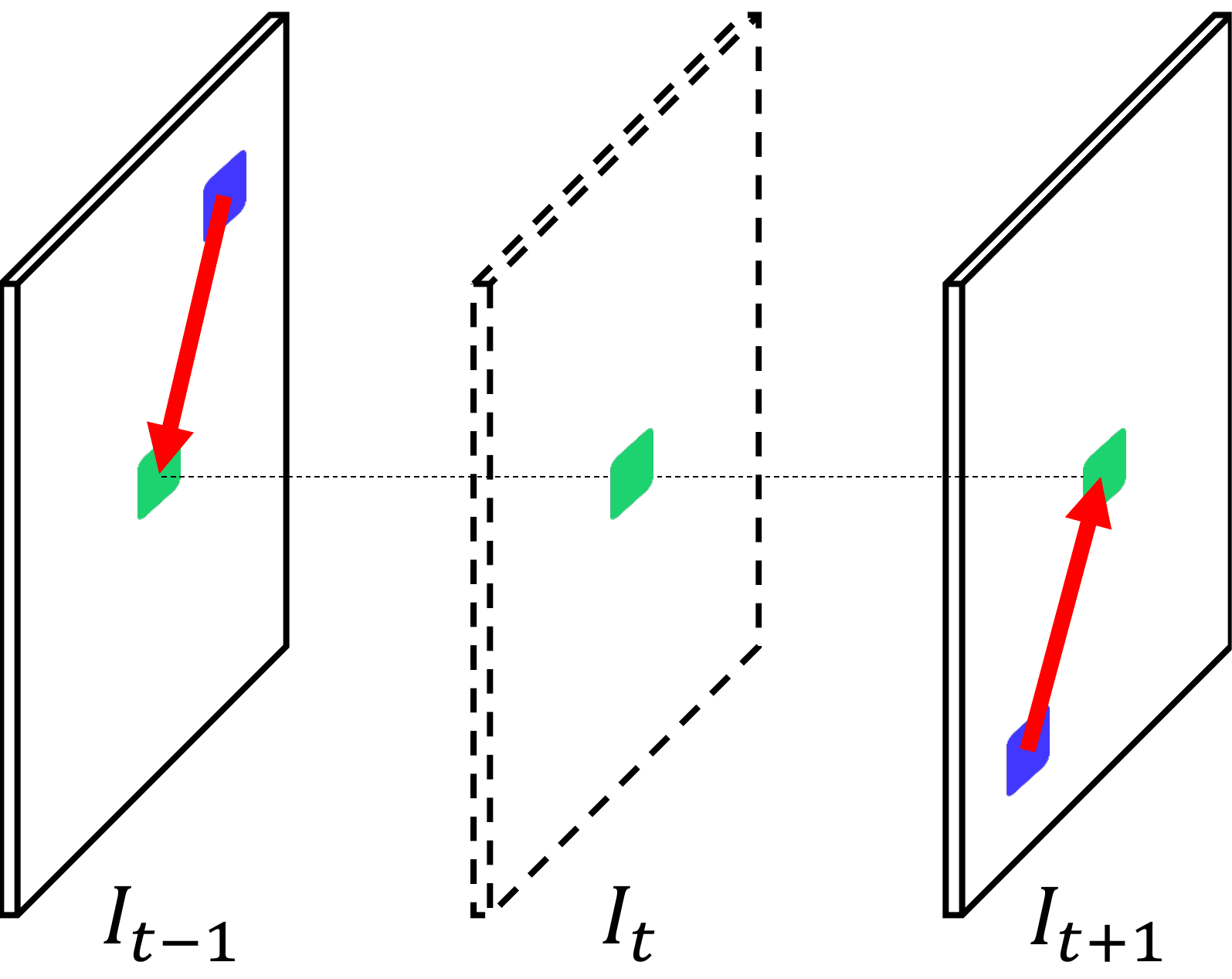}}\,
		\hfill
		\\[-2ex]
		\subfloat[Kernel and Flow Combined]
		{\includegraphics[width=0.46\linewidth]{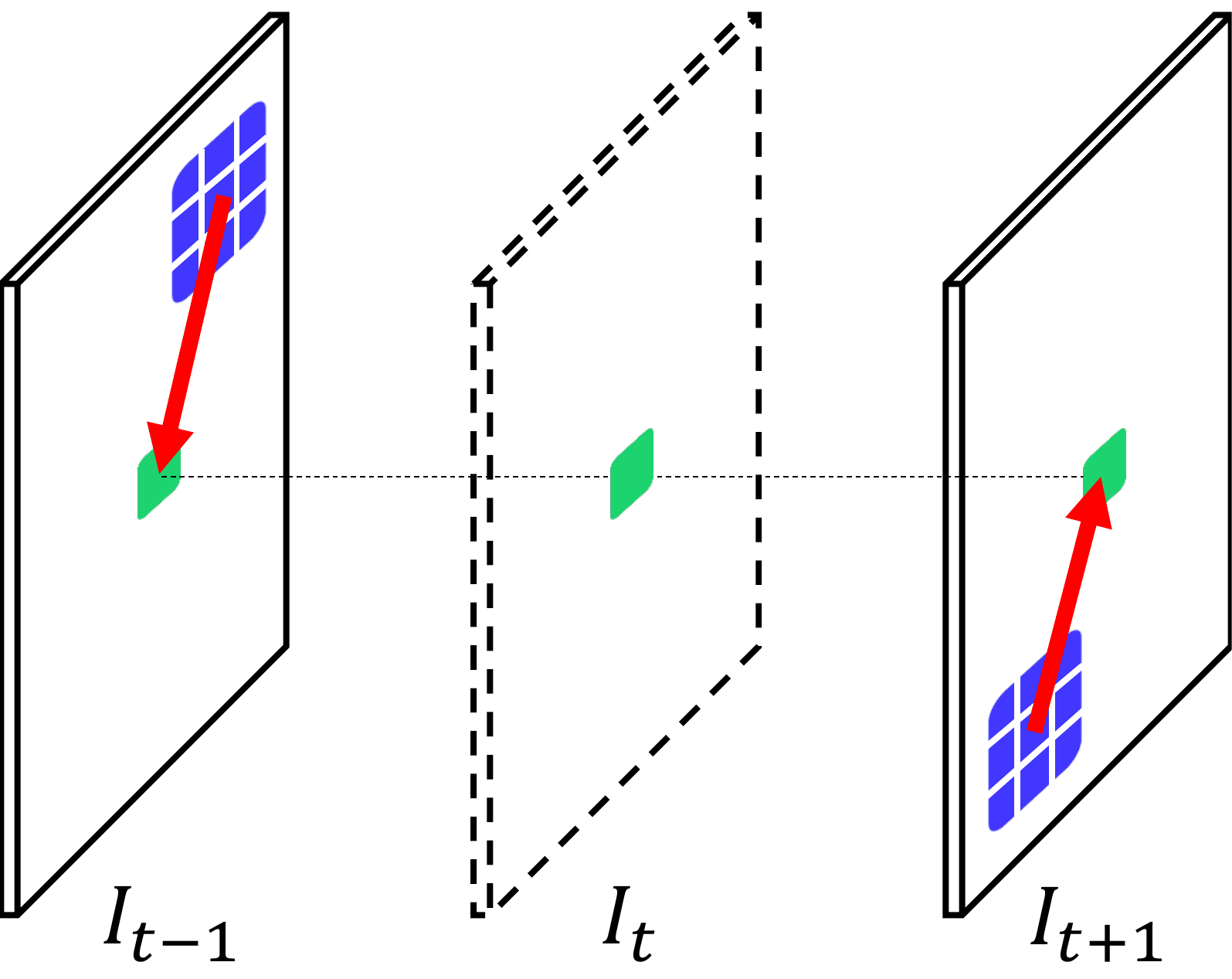}}\,
		\hfill
		\subfloat[Ours]
		{\includegraphics[width=0.46\linewidth]{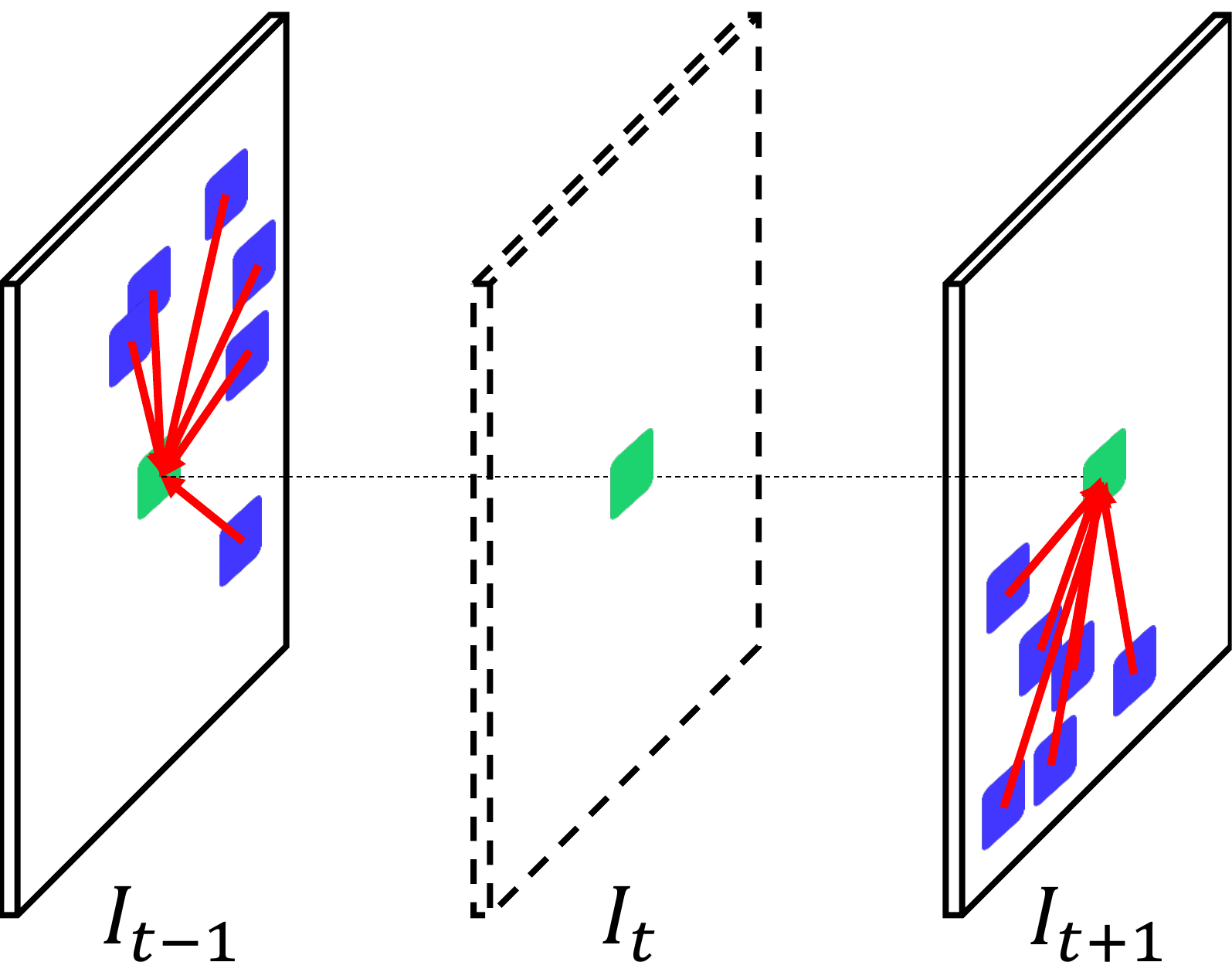}}\
		\hfill
		\caption{Overall description of the main streams and our method. The blue parts of each figure represent the reference points for generating the target pixel.}
		\label{fig:basic}
	\end{center}
\end{figure}

Most of the approaches define video frame interpolation as a problem of finding reference locations in input frames which include information for estimating each output pixel value. This can be seen as a motion estimation process, because the task involves tracking the path of the target pixel. Therefore, each algorithm covers its own motion domain, and this area is directly related to the performance. To handle motion in real world videos, we need a generalized operation that can refer to \emph{any number} of pixels in \emph{any location} in the input frames. However, most of the existing approaches have a variety of limitations in Degrees of Freedom~(DoF).\\
\indent One is the kernel-based approach~(Figure~\ref{fig:basic}~(a))~\cite{adaconv, sepconv}, which adaptively estimates the large-sized kernel for each pixel and synthesizes the intermediate frame by convolving the kernels with the input. This approach finds the proper reference location by assigning large weights to the pixels of interest. However, it does not refer to \emph{any location}, as it cannot deal with large motions beyond the kernel size. It is not efficient to keep the large size of the kernel even though the motion is small. The second approach is the flow-based approach~(Figure~\ref{fig:basic}~(b))~\cite{superslomo, deepvoxelflow}, which estimates the flow vector directly pointing to the reference location for each output pixel. However, it cannot refer to \emph{any number} of pixels because only one location is referred to in each input frame. Therefore, it is not suitable for complex motions and the result may suffer from lack of information when the input frame is of low-quality. Recently, methods of combining kernel-based and flow-based approaches are proposed to compensate for each other's limitations~(Figure~\ref{fig:basic}~(c))~\cite{xue2019video, bao2019memc}. They multiply the kernels with the location pointed to by the flow vector. Therefore, they can refer to \emph{any location} plus some additional neighboring pixels. However, this approach is not much different from the flow based approach as it uses significantly fewer reference points than the kernel-based one. In addition, there is room for improvement in terms of DoF because the shape of the kernel is a fixed square.\\
\indent In this paper, we propose an operation that refers to \emph{any number} of pixels and \emph{any location} called Adaptive Collaboration of Flows~(AdaCoF). To synthesize a target pixel, we estimate multiple flows, called offset vectors, pointing to the reference locations and sample them. Then the target pixel is obtained by linearly combining the sampled values. Our method is inspired by deformable convolution~(DefConv)~\cite{dai2017deformable}, but AdaCoF is significantly different from it in some points. First, DefConv has a shared weight for all positions, and it is not suitable for video because there are various motions in each position of a frame. Therefore, we allow the weights to be spatially adaptive. Second, AdaCoF is used as an independent module for frame warping, not for feature extraction as DefConv. Therefore, we obtain the weights as the outputs of a neural network, instead of training them as learnable parameters. Third, we add dilation for the starting point of the offset vectors to enforce the them to search a wider area. Lastly, we add an occlusion mask to utilize only one of the two input frames when one of the reference pixels is occluded. As shown in Figure~\ref{fig:basic}~(d), it can refer to \emph{any number} within \emph{any location} in the input frames, because the sizes and shapes of the kernels are not fixed. Therefore, our method has the highest DoF compared to most of the other competitive algorithms, and therefore can deal with various complex motions in real world videos. To make the synthesized frames more realistic, we further train a discriminator to detect the generated frame given the output and one of the input frames. Then we train the generator to maximize the entropy of the discriminator using dual-frame adversarial loss. Experimental results on various benchmarks show the effectiveness of AdaCoF over the latest state-of-the-art approaches.

\section{Related Work}
Most of the classic video frame interpolation methods estimate the dense flow maps using optical flow algorithms~\cite{dosovitskiy2015flownet, ilg2017flownet, sun2018pwc, weinzaepfel2013deepflow} and warp the input frames \cite{baker2011database, barron1994performance, werlberger2011optical, yu2013multi}. Therefore, the performance of these approaches largely depends on optical flow algorithms. Also, optical flow based approaches have limitations in many cases, such as occlusions, large motion, and brightness changes. Although there are some approaches without using external optical flow modules~\cite{liu2012multiple,mahajan2009moving}, they still have difficulty in dealing with these problems. Meyer~\emph{et~al.}~\cite{meyer2015phase} regard video frames as linear combinations of wavelets with different directions and frequencies. This approach interpolates each wavelet's phase and magnitude. This method makes notable progress in both performance and running time. Their recent work also applies deep learning to this approach \cite{meyer2018phasenet}. However, it still has limitations for large motions of high frequency components.\\
\indent Recent work has demonstrated the success of applying deep learning in the field of computer vision \cite{dong2016image, gatys2016image, he2016deep, johnson2016perceptual, krizhevsky2012imagenet, simonyan2014very}, which, in turn, inspires various deep learning based frame interpolation methods. As all we require for training neural networks are three consecutive video frames, learning based approaches are appropriate for this task. Long~\emph{et~al.}~\cite{long2016learning} propose a CNN architecture that uses two input frames and directly estimates the intermediate frame. However, this type of approach often leads to blurry results. Some other methods focus on where to find the output pixel from the input frames, instead of directly estimating the image. This paradigm is based on the fact that at least one input frame contains the output pixel, even in the case of occlusion. Niklaus~\emph{et~al.}~\cite{adaconv} estimate a kernel for each location and obtains the output pixel by convolving it over input patches. Each kernel samples the proper input pixels by combining them selectively. However, this requires a lot of memory and estimating large kernels for every pixel is computationally expensive. Niklaus~\emph{et~al.}~\cite{sepconv} solve this problem by estimating each kernel from the outer product of two vectors. However, this approach cannot handle motions larger than the kernel size and it is still wasteful to estimate large kernels for small motions. Liu~\emph{et~al.}~\cite{deepvoxelflow}
estimate a flow map that consists of vectors directly pointing to reference locations. They sample the proper pixels according to the flow map. However, as they assume that the forward and backward flows are the same, it is difficult to handle complex motions. Jiang~\emph{et~al.}~\cite{superslomo} propose a similar algorithm, but they estimate the forward and backward flows separately. They also improve the flow computation stage by defining the warping loss. However, it could be risky to get only one pixel value from each frame, especially when the input patches are of poor quality. To solve these problems, Reda~\emph{et~al.}~\cite{reda2018sdc} and Bao~\emph{et~al.}~\cite{bao2019memc} combine kernel and flow map based approaches. They multiply small-sized kernels with the locations pointed by the flow vectors. However, the reference points are still limited in a small area because the kernels maintain their square shape, which results in low DoF. 

There are some approaches that use additional information to solve problems in video frame interpolation. Niklaus~\emph{et~al.}~\cite{Niklaus_2018_CVPR} exploit the context informations extracted from ResNet-18~\cite{he2016deep} to enable the informative interpolation and succeed in obtaining high-quality results. In addition, Bao~\emph{et~al.}~\cite{bao2019depth} use depth maps estimated from hourglass architecture~\cite{chen2016single} to solve the occlusion problems. Lastly, Liu~\emph{et~al.}~\cite{liu2019deep} obtain better performance with cycle consistency loss and additional edge maps. These approaches can be independently applied to many other algorithms, including our approach.

\section{Proposed Approach}
\subsection{Video Frame Interpolation}
\label{redefine}
Given consecutive video frames $I_n$ and $I_{n+1}$, where $n \in \mathbb{Z}$ is a frame index, our goal is to find the intermediate frame $I_{out}$. All the information required to produce $I_{out}$ can be obtained from $I_n$ and $I_{n+1}$. Therefore, all we have to do is find the relations between them. We regard the relation as a warping operation $\mathcal{T}$ from $I_n$ and $I_{n+1}$ to $I_{out}$. For the forward and backward warping operations $\mathcal{T}_{f}$ and $\mathcal{T}_{b}$, we can consider $I_{out}$ as a combination of $\mathcal{T}_{f}(I_n)$ and $\mathcal{T}_{b}(I_{n+1})$ as follows.

\begin{equation}
I_{out} = \mathcal{T}_{f}(I_n) + \mathcal{T}_{b}(I_{n+1})
\label{eq:base}
\end{equation}

\noindent The frame interpolation task results in a problem of how the spatial transform $\mathcal{T}$ can be found. We employ a new operation called \textit{Adaptive Collaboration of Flows~(AdaCoF)} for $\mathcal{T}$, which convolve the input image with adaptive kernel weights and offset vectors for each output pixel.

\noindent\textbf{Occlusion reasoning.} Let both the input and output image sizes be $M \times N$. In the case of occlusion, the target pixel will not be visible in one of the input images. Therefore we define occlusion map $V \in [0,1]^{M \times N}$ and modify Equation (\ref{eq:base}) as follows.

\begin{equation}
I_{out} = V \odot \mathcal{T}_{f}(I_n) + (J-V) \odot \mathcal{T}_{b}(I_{n+1}),
\end{equation}

\noindent where $\odot$ is a pixel-wise multiplication and $J$ is an $M \times N$ matrix of ones. For the target pixel $(i,j)$, $V(i,j)=1$ implies that the pixel is visible only in $I_n$ and $V(i,j)=0$ implies that it is visible only in $I_{n+1}$.

\begin{figure}
	\setlength{\belowcaptionskip}{-20pt}
	\begin{center}
		\subfloat[$d=0$]
		{\includegraphics[width=0.32\linewidth]{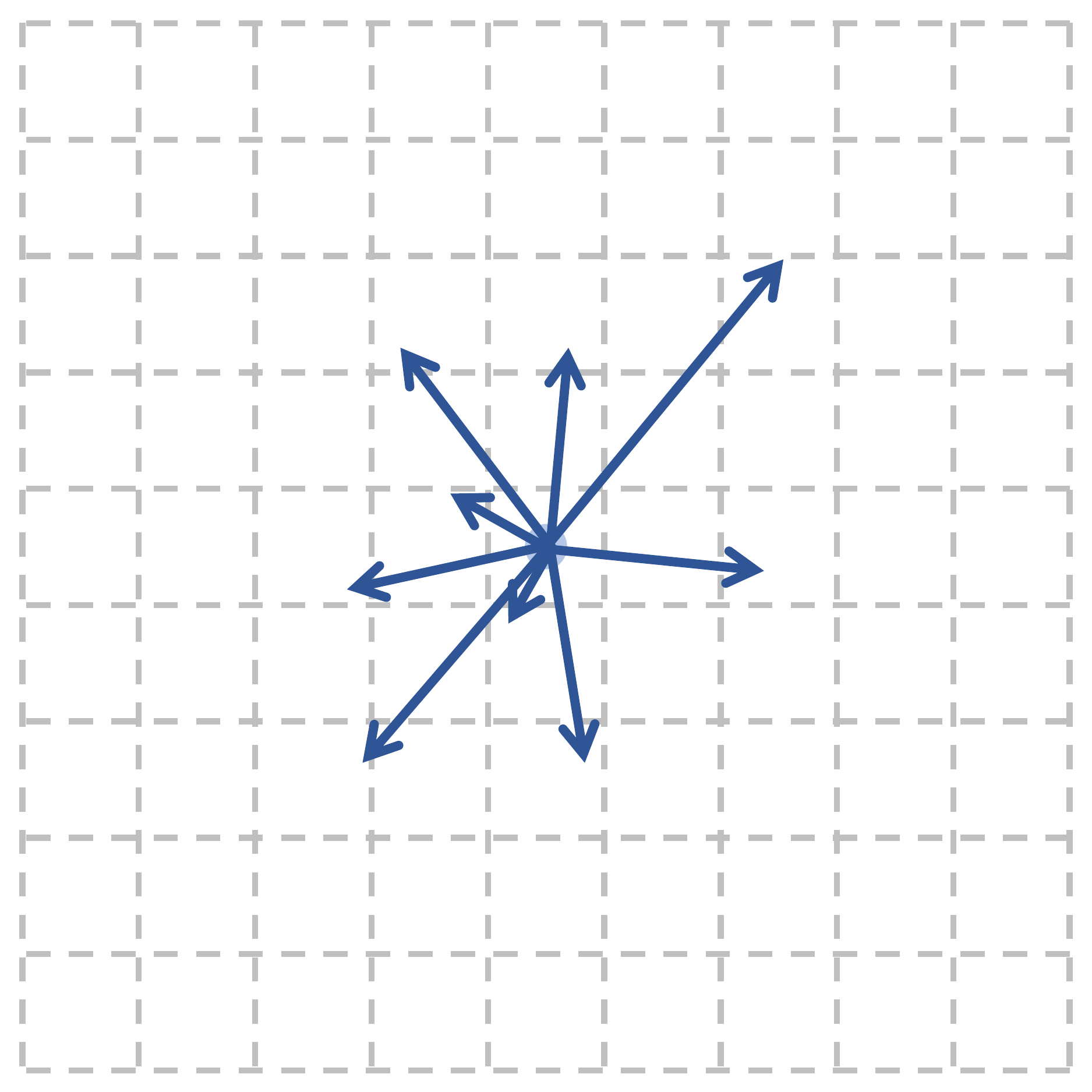}}\,
		\subfloat[$d=1$]
		{\includegraphics[width=0.32\linewidth]{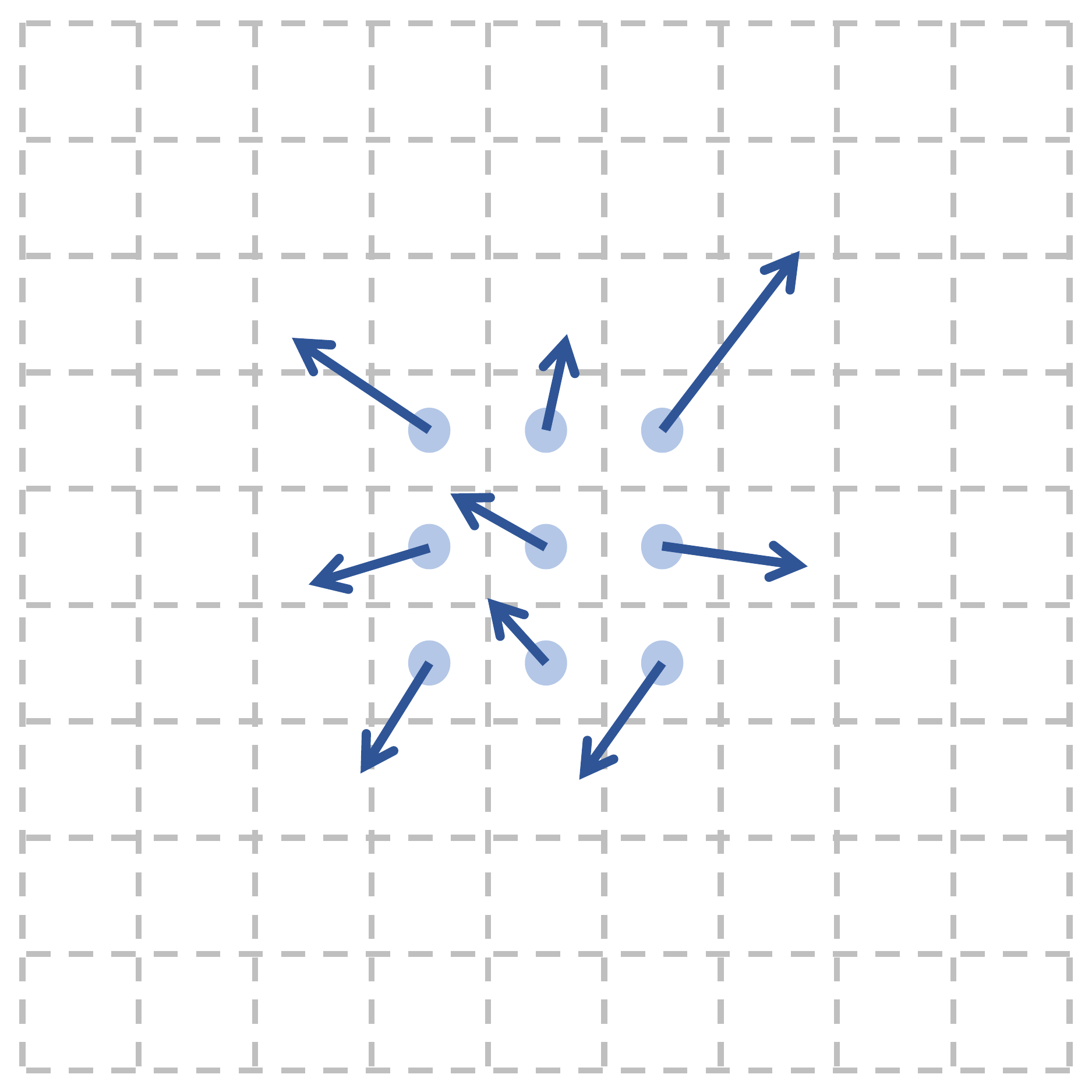}}\,
		\subfloat[$d=2$]
		{\includegraphics[width=0.32\linewidth]{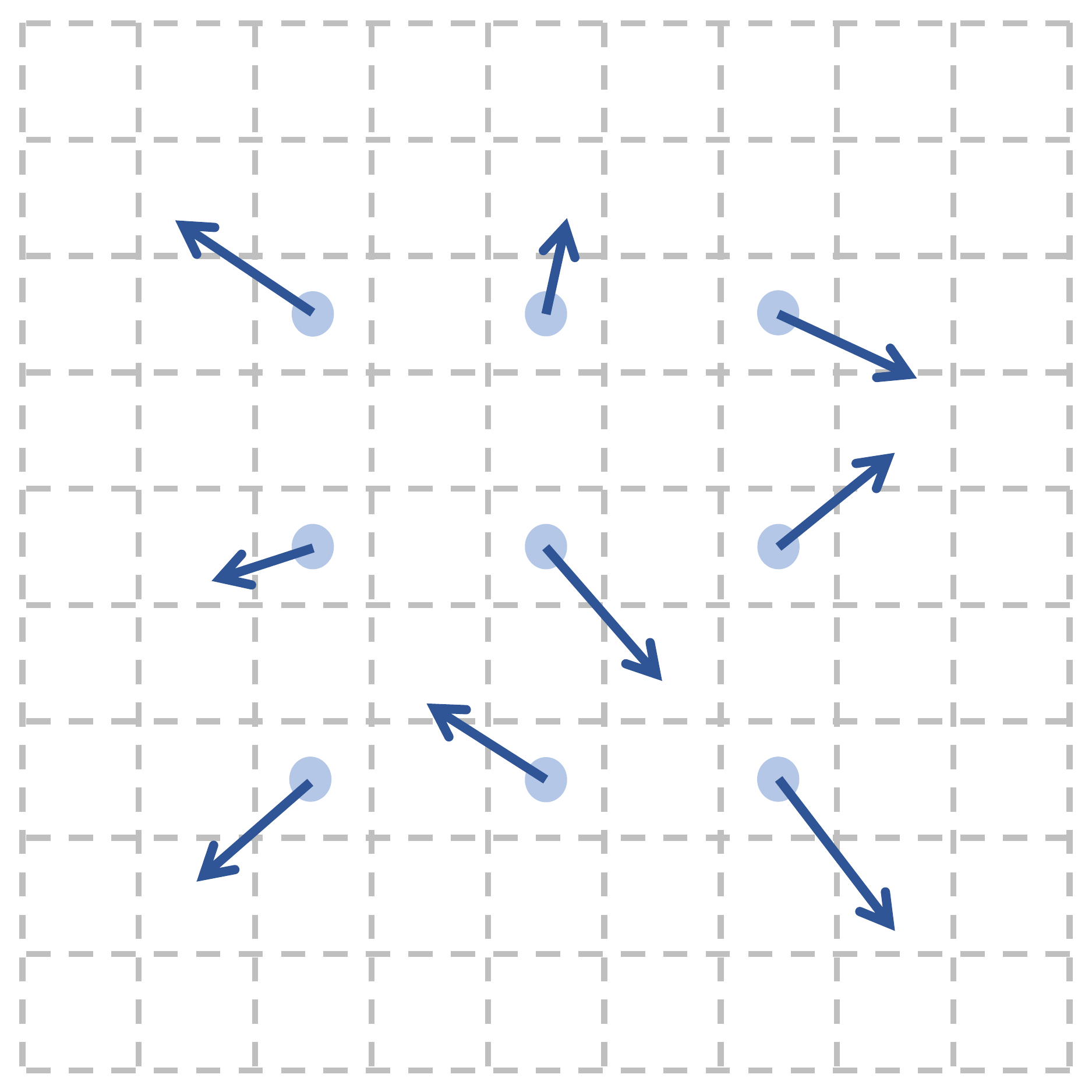}}\
		
		\caption{Illustration of the offset vectors of AdaCoF under various dilations.}
		\label{fig:dilation}
	\end{center}
\end{figure}

\begin{figure*}
	\setlength{\belowcaptionskip}{-10pt}
	\begin{center}
		\includegraphics[width=0.9\linewidth]{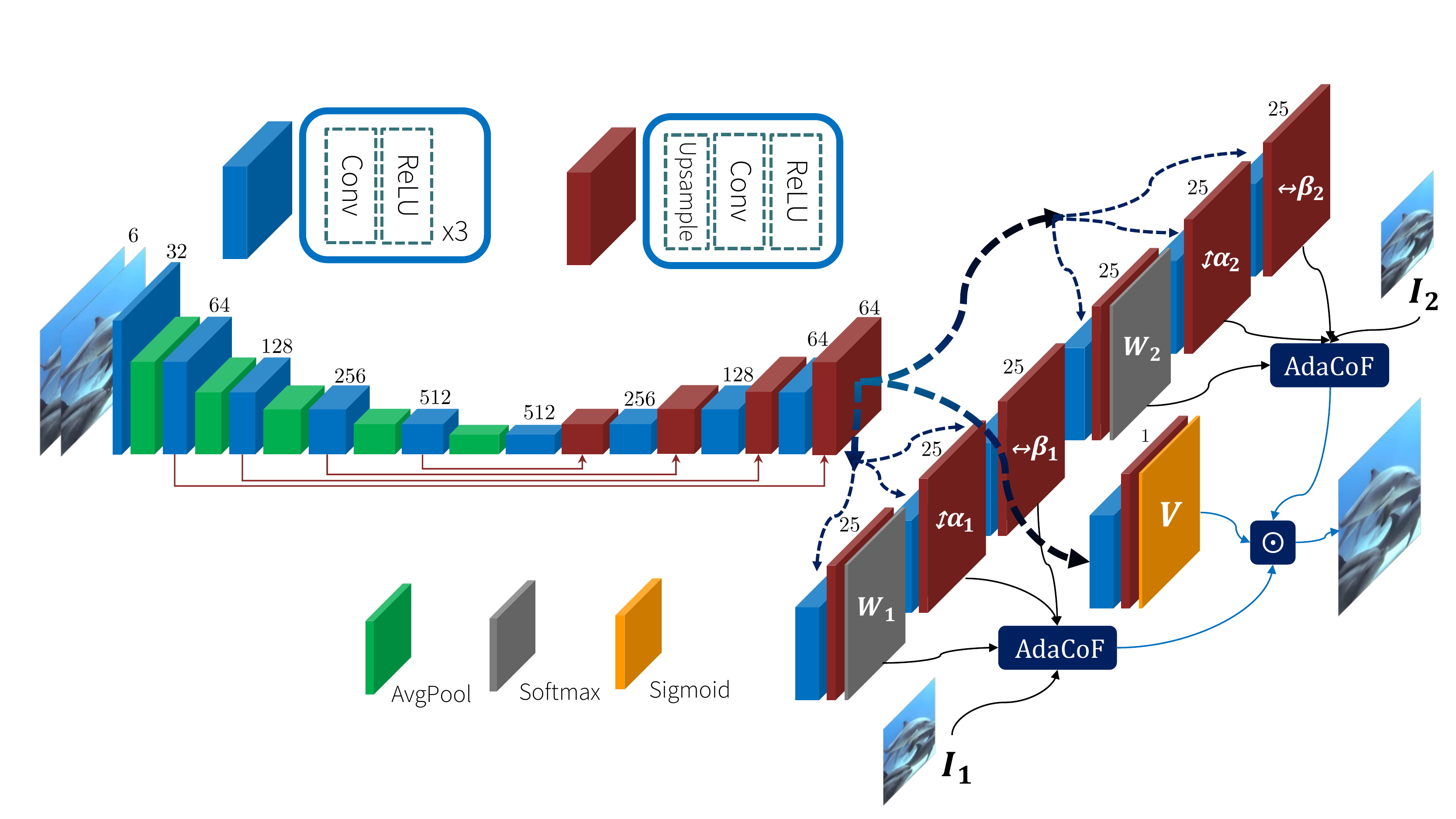}
	\end{center}
	\vspace{-7.5mm}
	\caption{The neural network architecture. The model consists of three main parts: the U-Net, sub-networks, and Adaptive Collaboration of Flows~(AdaCoF). The U-Net architecture extracts features from the input image. Then the sub-networks estimates the parameters needed for AdaCoF from the extracted features. The output's height and width of each sub-network are the same as that of the input. Each parameter group for an output pixel is obtained as a 1D vector along the channel axis. The AdaCoF part synthesizes the intermediate frame using the input frames and parameters.}
	\label{fig:network}
\end{figure*}

\subsection{Adaptive Collaboration of Flows}
\label{defconv}
Let the frame warped from $I$ be $\hat{I}$. When we define $\mathcal{T}$ as a classic convolution, we can write $\hat{I}$ as follows.
\begin{equation}
\hat{I}(i,j) = \sum_{k=0}^{F-1}{\sum_{l=0}^{F-1}{W_{k,l}I(i+k,j+l)}},
\end{equation}
\noindent where $F$ is the kernel size and $W_{k,l}$ are the kernel weights. The input image $I$ is considered to be padded so that the original input and output size are equal. Deformable convolution \cite{dai2017deformable} adds offset vectors $\Delta p_{k,l} = (\alpha_{k,l}, \beta_{k,l})$ to the classic convolution as follows.
\begin{equation}
\hat{I}(i,j) = \sum_{k=0}^{F-1}{\sum_{l=0}^{F-1}{W_{k,l}I(i+k+\alpha_{k,l},j+l+\beta_{k,l})}}
\end{equation}
\noindent AdaCoF, unlike the classic deformable convolutions, does not share the kernel weights over the different pixels. Therefore the notation for the kernel weights $W_{k,l}$ should be written as follows.
\begin{equation}
\hat{I}(i,j) = \sum_{k=0}^{F-1}{\sum_{l=0}^{F-1}{W_{k,l}(i,j)I(i+k+\alpha_{k,l},j+l+\beta_{k,l})}}
\end{equation}

\noindent The offset values $\alpha_{k,l}$ and $\beta_{k,l}$ may not be integer values. In other words, $(\alpha_{k,l}, \beta_{k,l})$ could point to an arbitrary location, not only the grid point. Therefore, the pixel value of $I$ for any location has to be defined. We use bilinear interpolation to obtain the values of non-grid location as DCNs~\cite{dai2017deformable}. It also makes the module differentiable; therefore, the whole network can be trained end-to-end.

\noindent\textbf{Dilation.} We found that dilating the starting point of the offset vectors helps AdaCoF to explore wider area as shown in Figure~\ref{fig:dilation}. Therefore, we add dilation term $d \in \{ 0, 1, 2, ...\}$ to the operation as follows.

\begin{equation}
\begin{multlined}
\hat{I}(i,j) =\\
\sum_{k=0}^{F-1}{\sum_{l=0}^{F-1}{W_{k,l}(i,j)I(i+dk+\alpha_{k,l},j+dl+\beta_{k,l})}}
\end{multlined}
\end{equation}

\subsection{Network Architecture}
\label{network}

We design a fully convolutional neural network which estimates the kernel weights $W_{k,l}$, offset vectors $(\alpha_{k,l}, \beta_{k,l})$, and occlusion map $V$. Therefore, any video frames size can be used as the input. Furthermore, because each module of the neural network is differentiable, it is end-to-end trainable. Our neural network starts with the U-Net architecture, which consists of encoder, decoder, and skip connections~\cite{10.1007/978-3-319-24574-4_28}. Each processing unit basically contains 3~$\times$~3 convolution and ReLU activation. For the encoder part, we use average pooling to extract the features. And for the decoder part, we use bilinear interpolation for the upsampling. After the U-Net architecture, the seven sub-networks finally estimate the outputs ($W_{k,l}$, $\alpha_{k,l}$, $\beta_{k,l}$ for each frame and $V$). We use sigmoid activation for $V$ to satisfy $V \in [0,1]^{M \times N}$. Moreover, as the weights $W_{k,l}$ for each pixel have to be non-negative and must add up to 1, softmax layers are used for the constraints. More specific architectures of the network are described in Figure~\ref{fig:network}.

\subsection{Objective Functions}
\label{objective}

\noindent\textbf{Loss Function.} First, we have to reduce a difference between the model output $I_{out}$ and ground truth $I_{gt}$. We use $\ell_1$ norm for the loss as follows.

\begin{equation}
\mathcal{L}_1 = \|I_{out} - I_{gt}\|_1
\end{equation}

\noindent The $\ell_2$ norm can be used, but it is known that the $\ell_2$~norm-based optimization leads to blurry results in most of the image synthesis tasks \cite{NIPS2015_5951, long2016learning, mathieu2015deep, srivastava2015unsupervised}. Following Liu~\emph{et~al.}~\cite{deepvoxelflow}, we use the Charbonnier Function $\Phi(x)=(x^2 + \epsilon^2)^{1/2}$ for optimizing $\ell_1$ norm, where $\epsilon=0.001$.

\noindent\textbf{Perceptual Loss.} Perceptual loss has been found to be effective in producing visually more realistic outputs~\cite{dosovitskiy2016generating, johnson2016perceptual, zhu2016generative}. We add the perceptual loss with the feature extractor $\mathcal{F}$ from \texttt{conv4\textunderscore3} of ImageNet pretrained VGG16 network.

\begin{equation}
\mathcal{L}_{vgg} = \|\mathcal{F}(I_{out}) - \mathcal{F}(I_{gt})\|_2
\end{equation}

\noindent\textbf{Dual-Frame Adversarial Loss.} It is known that training the networks with adversarial loss~\cite{goodfellow2014generative} can lead to results of higher quality and sharpness, instead of increasing mean squared error~\cite{ledig2017photo, blau2018perception}. This could be applied to video frame interpolation tasks. However, simply applying it to the single output frame does not consider the temporal consistency and leads to a disparate result compared to the input frames. What we want is to make the synthesized frame appear natural among the adjacent frames, not the other real images. Therefore, we concatenate the generated frame and one of the input frames in the temporal order and train the discriminator $C$ to distinguish which of the two is the generated frame with the following loss.

\begin{equation}
-\mathcal{L}_C = \log(C(\left[I_n, I_{out}\right]))+\log(1-C(\left[I_{out}, I_{n+1}\right])),
\end{equation}

\noindent where $\left[ \cdot \right]$ is concatenation. Then we train the main network to maximize the uncertainty, i.e., entropy, of the discriminator with the following loss. This idea is inspired by some prior works \cite{denton2017unsupervised,ganin2016domain}.

\begin{equation}
\begin{aligned}
\mathcal{L}_{adv} =& C(\left[I_n, I_{out}\right])\log(C(\left[I_n, I_{out}\right]))\\
&+C(\left[I_{out}, I_{n+1}\right])\log(C(\left[I_{out}, I_{n+1}\right]))
\end{aligned}
\end{equation}

\begin{table}
	\setlength{\belowcaptionskip}{-20pt}
	\begin{center}
		\resizebox{\columnwidth}{!}{
			\begin{tabular}{lcccccc}
				\toprule
				& \multicolumn{2}{c}{Middlebury} & \multicolumn{2}{c}{UCF101} & \multicolumn{2}{c}{DAVIS} \\
				\cmidrule(l{5pt}r{5pt}){2-3} \cmidrule(l{5pt}r{5pt}){4-5} \cmidrule(l{5pt}r{5pt}){6-7}
				& PSNR  & SSIM  & PSNR  & SSIM  & PSNR  & SSIM \\
				\midrule
				Ours-\emph{fb} & 32.879  & 0.956  & 33.449  & 0.967  & 24.787  & 0.828  \\
				Ours-\emph{kb} & 34.762  & 0.972  & 34.689  & 0.973  & 25.802  & 0.854  \\
				Ours-\emph{ws} & 35.412  & 0.976  & 34.901  & 0.973  & 26.623  & 0.866  \\
				Ours-\emph{woocc} & 35.471  & 0.975  & 34.907  & 0.973  & 26.482  & 0.863  \\
				Ours-\emph{sdc} & 34.973  & 0.972  & 34.673  & \textbf{\textcolor{red}{0.974}}  & 26.367  & 0.866  \\
				Ours-\emph{vgg} & \textbf{\textcolor{blue}{35.694}}  & \textbf{\textcolor{blue}{0.977}}  & \textbf{\textcolor{blue}{34.973}}  & 0.973  & \textbf{\textcolor{red}{26.773}}  & \textbf{\textcolor{red}{0.869}}  \\
				Ours & \textbf{\textcolor{red}{35.715}}  & \textbf{\textcolor{red}{0.978}}  & \textbf{\textcolor{red}{35.063}}  & \textbf{\textcolor{red}{0.974}}  & \textbf{\textcolor{blue}{26.636}}  & \textbf{\textcolor{blue}{0.868}}  \\
				\bottomrule
		\end{tabular}}%
		\vspace{-1mm}
		\caption{Result of ablation study on warping operations.}
		\label{tbl:ablation}
	\end{center}
\end{table}

\noindent Thus, the network is intended to generate an output that is realistic compared to the adjacent input frames.\\

We finally combine above losses to compose two versions of objective function: distortion-oriented loss~($\mathcal{L}_d$) and perception-oriented loss~($\mathcal{L}_p$) as follows.

\begin{equation}
\mathcal{L}_d = \mathcal{L}_1,
\end{equation}
\begin{equation}
\mathcal{L}_p = \lambda_1 \mathcal{L}_1 + \lambda_{vgg} \mathcal{L}_{vgg} + \lambda_{adv} \mathcal{L}_{adv},
\end{equation}

\noindent For the perception-oriented version, we first train the network with $\mathcal{L}_d$ then fine-tune it with $\mathcal{L}_p$.

\section{Experiments}
\subsection{Experimental Settings}
\label{implementation}
\noindent\textbf{Learning Strategy.} We train our neural network using AdaMax optimizer~\cite{kingma2014adam}, where $\beta_1 = 0.9, \beta_2 = 0.999$. The learning rate is initially 0.001 and decays half every 20 epochs. The batch size is 4 and the network is trained for 50 epochs.

\noindent\textbf{Training Dataset.} We use Vimeo90K~\cite{xue2019video} dataset for training. It contains 51,312 triplets of $256 \times 448$ video frames. To augment the dataset, we randomly crop $256 \times 256$ patches from the original images. We also eliminate the biases due to the priors by flipping horizontally, vertically and swapping the order of frames for the probability 0.5.

\begin{table}
	\setlength{\belowcaptionskip}{-10pt}
	\begin{center}
		\resizebox{\columnwidth}{!}{
			\begin{tabular}{lcccccc}
				\toprule
				& \multicolumn{2}{c}{Middlebury} & \multicolumn{2}{c}{UCF101} & \multicolumn{2}{c}{DAVIS} \\
				\cmidrule(l{5pt}r{5pt}){2-3} \cmidrule(l{5pt}r{5pt}){4-5} \cmidrule(l{5pt}r{5pt}){6-7}
				& PSNR  & SSIM  & PSNR  & SSIM  & PSNR  & SSIM \\
				\midrule
				$F=1$ & 32.879  & 0.956  & 33.449  & 0.967  & 24.787  & 0.828  \\
				$F=3$ & 35.212  & 0.975  & 34.728  & 0.973  & 26.535  & 0.867  \\
				$F=5$ & 35.715  & 0.978  & \textbf{\textcolor{red}{35.063}}  & \textbf{\textcolor{red}{0.974}}  & 26.636  & 0.868  \\
				$F=7$ & 35.927  & 0.979  & 34.974  & \textbf{\textcolor{red}{0.974}}  & \textbf{\textcolor{blue}{26.987}}  & \textbf{\textcolor{blue}{0.873}}  \\
				$F=9$ & \textbf{\textcolor{blue}{36.019}}  & \textbf{\textcolor{blue}{0.980}}  & 35.012  & 0.973  & \textbf{\textcolor{red}{27.029}}  & \textbf{\textcolor{red}{0.875}}  \\
				$F=11$ & \textbf{\textcolor{red}{36.094}}  & \textbf{\textcolor{red}{0.981}}  & \textbf{\textcolor{blue}{35.024}}  & \textbf{\textcolor{red}{0.974}}  & 26.941  & \textbf{\textcolor{blue}{0.873}}  \\
				\bottomrule
		\end{tabular}}%
		\vspace{-1mm}
		\caption{Experimental result on kernel size $F$.}
		\label{tbl:kernelsize}
	\end{center}
\end{table}

\begin{table}
	\setlength{\belowcaptionskip}{-20pt}
	\begin{center}
		\resizebox{\columnwidth}{!}{
			\begin{tabular}{lcccccc}
				\toprule
				& \multicolumn{2}{c}{Middlebury} & \multicolumn{2}{c}{UCF101} & \multicolumn{2}{c}{DAVIS} \\
				\cmidrule(l{5pt}r{5pt}){2-3} \cmidrule(l{5pt}r{5pt}){4-5} \cmidrule(l{5pt}r{5pt}){6-7}
				& PSNR  & SSIM  & PSNR  & SSIM  & PSNR  & SSIM \\
				\midrule
				$d=0$ & 35.489  & 0.977  & 35.032  & \textbf{\textcolor{blue}{0.974}}  & \textbf{\textcolor{blue}{26.710}}  & \textbf{\textcolor{red}{0.870}}  \\
				$d=1$ & \textbf{\textcolor{blue}{35.715}}  & \textbf{\textcolor{blue}{0.978}}  & \textbf{\textcolor{blue}{35.063}}  & \textbf{\textcolor{blue}{0.974}}  & 26.636  & 0.868  \\
				$d=2$ & \textbf{\textcolor{red}{35.876}}  & \textbf{\textcolor{red}{0.980}}  & \textbf{\textcolor{red}{35.099}}  & \textbf{\textcolor{red}{0.974}}  & \textbf{\textcolor{red}{26.910}}  & \textbf{\textcolor{red}{0.870}}  \\
				\bottomrule
		\end{tabular}}%
		\vspace{-1mm}
		\caption{Experimental result on dilation $d$.}
		\label{tbl:dilation}
	\end{center}
\end{table}

\begin{table*}
	\Large
	\setlength{\belowcaptionskip}{-15pt}
	\begin{center}
		\resizebox{\textwidth}{!}{
			\begin{tabular}{lcc|ccccccccccccccccccccccccc}
				\toprule
				& \multicolumn{2}{c}{\sc Average} & \multicolumn{2}{c}{Mequon} & \multicolumn{2}{c}{Schefflera} & \multicolumn{2}{c}{Urban} & \multicolumn{2}{c}{Teddy} & \multicolumn{2}{c}{Backyard} & \multicolumn{2}{c}{Basketball} & \multicolumn{2}{c}{Dumptruck} & \multicolumn{2}{c}{Evergreen}\\
				\cmidrule(l{5pt}r{5pt}){2-3} \cmidrule(l{5pt}r{5pt}){4-5} \cmidrule(l{5pt}r{5pt}){6-7} \cmidrule(l{5pt}r{5pt}){8-9} \cmidrule(l{5pt}r{5pt}){10-11} \cmidrule(l{5pt}r{5pt}){12-13} \cmidrule(l{5pt}r{5pt}){14-15} \cmidrule(l{5pt}r{5pt}){16-17} \cmidrule(l{5pt}r{5pt}){18-19}
				& IE  & NIE  & IE  & NIE & IE  & NIE & IE  & NIE & IE  & NIE & IE  & NIE & IE  & NIE & IE  & NIE & IE  & NIE \\
				\midrule
				MDP-Flow2~\cite{xu2012motion} & 5.83  & 0.87  & 2.89  & 0.59  & 3.47  & 0.62  & 3.66  & 1.24  & 5.20  & 0.94  & 10.20  & 0.98  & 6.13  & 1.09  & 7.36  & 0.70  & 7.75  & 0.78  \\
				
				DeepFlow~\cite{weinzaepfel2013deepflow} & 5.97  & 0.86  & 2.98  & 0.62  & 3.88  & 0.74  & 3.62  & 0.86  & 5.39  & 0.99  & 11.00  & 1.04  & 5.91  & 1.02  & 7.14  & 0.63  & 7.80  & 0.96  \\
				
				SepConv~\cite{sepconv} & 5.61  & 0.83  & 2.52  & \textbf{\textcolor{blue}{0.54}}  & 3.56  & 0.67  & 4.17  & 1.07  & 5.41  & 1.03  & 10.20  & 0.99  & 5.47  & 0.96  & 6.88  & 0.68  & 6.63  & 0.70  \\
				
				SuperSlomo~\cite{superslomo} & 5.31  & 0.78  & 2.51  & 0.59  & 3.66  & 0.72  & \textbf{\textcolor{blue}{2.91}}  & 0.74  & 5.05  & 0.98  & 9.56  & 0.94  & 5.37  & 0.96  & 6.69  & 0.60  & 6.73  & 0.69  \\
				
				CtxSyn~\cite{Niklaus_2018_CVPR} & 5.28  & 0.82  & \textbf{\textcolor{red}{2.24}}  & \textbf{\textcolor{red}{0.50}}  & \textbf{\textcolor{red}{2.96}}  & \textbf{\textcolor{red}{0.55}}  & 4.32  & 1.42  & \textbf{\textcolor{red}{4.21}}  & \textbf{\textcolor{blue}{0.87}}  & 9.59  & 0.95  & 5.22  & 0.94  & 7.02  & 0.68  & 6.66  & 0.67  \\
				
				CyclicGen~\cite{liu2019deep} & \textbf{\textcolor{red}{4.20}}  & \textbf{\textcolor{blue}{0.73}}  & \textbf{\textcolor{blue}{2.26}}  & 0.64  & 3.19  & 0.67  & \textbf{\textcolor{red}{2.76}}  & \textbf{\textcolor{blue}{0.72}}  & 4.97  & 0.95  & \textbf{\textcolor{blue}{8.00}}  & 0.91  & \textbf{\textcolor{red}{3.36}}  & 0.87  & \textbf{\textcolor{red}{4.55}}  & \textbf{\textcolor{red}{0.53}}  & \textbf{\textcolor{red}{4.48}}  & \textbf{\textcolor{red}{0.52}}  \\
				
				TOF-M~\cite{xue2019video} & 5.49  & 0.84  & 2.54  & 0.55  & 3.70  & 0.72  & 3.43  & 0.92  & 5.05  & 0.96  & 9.84  & 0.97  & 5.34  & 0.98  & 6.88  & 0.72  & 7.14  & 0.90  \\
				
				DAIN~\cite{bao2019depth} & 4.86  & \textbf{\textcolor{red}{0.71}}  & 2.38  & 0.58  & 3.28  & 0.60  & 3.32  & \textbf{\textcolor{red}{0.69}}  & \textbf{\textcolor{blue}{4.65}}  & \textbf{\textcolor{red}{0.86}}  & \textbf{\textcolor{red}{7.88}}  & \textbf{\textcolor{red}{0.87}}  & 4.73  & 0.85  & 6.36  & 0.59  & 6.25  & 0.66  \\
				
				MEMC-Net~\cite{bao2019memc} & 5.00  & 0.74  & 2.39  & 0.59  & 3.36  & 0.64  & 3.37  & 0.80  & 4.84  & 0.88  & 8.55  & \textbf{\textcolor{blue}{0.88}}  & 4.70  & 0.85  & 6.40  & 0.64  & 6.37  & 0.63  \\
				
				AdaCoF~(Ours) & \textbf{\textcolor{blue}{4.75}}  & \textbf{\textcolor{blue}{0.73}}  & 2.41  & 0.60  & \textbf{\textcolor{blue}{3.10}}  & \textbf{\textcolor{blue}{0.59}}  & 3.48  & 0.84  & 4.84  & 0.92  & 8.68  & 0.90  & \textbf{\textcolor{blue}{4.13}}  & \textbf{\textcolor{red}{0.84}}  & \textbf{\textcolor{blue}{5.77}}  & \textbf{\textcolor{blue}{0.58}}  & \textbf{\textcolor{blue}{5.60}}  & \textbf{\textcolor{blue}{0.57}}  \\
				\bottomrule
		\end{tabular}}%
		\caption{Evaluation results on the Middlebury benchmark.}
		\label{tbl:middlebury}
	\end{center}
\end{table*}

\begin{table}
	\setlength{\belowcaptionskip}{-15pt}
	\begin{center}
		\resizebox{\columnwidth}{!}{
			\begin{tabular}{lcccccc}
				\toprule
				& \multicolumn{2}{c}{Middlebury} & \multicolumn{2}{c}{UCF101} & \multicolumn{2}{c}{DAVIS} \\
				\cmidrule(l{5pt}r{5pt}){2-3} \cmidrule(l{5pt}r{5pt}){4-5} \cmidrule(l{5pt}r{5pt}){6-7}
				& PSNR  & SSIM  & PSNR  & SSIM  & PSNR  & SSIM \\
				\midrule
				Overlapping & 27.968  & 0.879  & 30.445  & 0.935  & 21.922  & 0.740  \\
				Phase Based~\cite{meyer2015phase} & 31.117  & 0.933  & 32.454  & 0.953  & 23.465  & 0.800  \\
				MIND~\cite{long2016learning}  & 31.346  & 0.943  & 32.437  & 0.963  & 25.570  & 0.852  \\
				SepConv~\cite{sepconv} & 35.521  & 0.977  & 34.735  & 0.973  & 26.258  & 0.861  \\
				DVF~\cite{deepvoxelflow}   & 34.340  & 0.971  & 34.465  & 0.972  & 25.880  & 0.858  \\
				SuperSlomo~\cite{superslomo} & 34.234  & 0.972  & 34.055  & 0.970  & 25.699  & 0.858  \\
				Ours & \textbf{\textcolor{blue}{35.715}} & \textbf{\textcolor{blue}{0.978}} & \textbf{\textcolor{red}{35.063}} & \textbf{\textcolor{red}{0.974}} & \textbf{\textcolor{blue}{26.636}} & \textbf{\textcolor{blue}{0.868}} \\
				Ours~+ & \textbf{\textcolor{red}{36.139}}  & \textbf{\textcolor{red}{0.981}}  & \textbf{\textcolor{blue}{35.048}}  & \textbf{\textcolor{red}{0.974}}  & \textbf{\textcolor{red}{27.070}}  & \textbf{\textcolor{red}{0.874}}  \\
				\bottomrule
		\end{tabular}}%
		\caption{Evaluation result with fixed train dataset.}
		\label{tbl:trainfix}
	\end{center}
\end{table}

\noindent\textbf{Computational issue.} Our approach is implemented using PyTorch~\cite{paszke2017automatic}. To implement the AdaCoF layer, we used CUDA and cuDNN~\cite{chetlur2014cudnn} for the parallel processing. We set the kernel size $5 \times 5$ and all the weights, offsets and occlusion map require 0.94 GB of memory for a 1080p video frame. It is about 70\% demand compared to Niklaus~\emph{et~al.}~\cite{sepconv}. Using RTX 2080 Ti GPU, it takes 0.21 seconds to synthesize a $1280 \times 720$ frame.

\noindent\textbf{Evaluation settings.} The test datasets used for the experiments are the Middlebury dataset~\cite{baker2011database}, some randomly sampled sequences from UCF101~\cite{soomro2012ucf101} and the DAVIS dataset~\cite{Perazzi2016}. We evaluate each algorithm by measuring PSNR~(Peak Signal-to-Noise Ratio) and SSIM~(Structural Similarity)~\cite{wang2004image} for all test datasets. For all the tables in this section, the \textbf{\textcolor{red}{red}} numbers mean the best performance and the \textbf{\textcolor{blue}{blue}} numbers mean the second best performance.

\subsection{Ablation Study}
\label{ablation}

We analyze the contributions of each module in terms of five keywords: warping operation, perceptual loss, kernel size, dilation and adversarial loss.

\noindent\textbf{Warping Operation.} To verify that higher DoF leads to better performance, we fixed the backbone network and replaced AdaCoF with some other warping operations of lower DoF. We train all versions of warping operation with $\mathcal{L}_d$ and the kernel sizes are fixed to be 5 except for Ours-\emph{fb}.
\begin{itemize}
	\setlength{\itemsep}{-3pt}
	\item Ours-\emph{fb}: To compare AdaCoF with flow-based approaches, we set the kernel size to be 1.
	\item Ours-\emph{kb}: SepConv~\cite{sepconv} is one of the most representative kernel-based approaches. However, because it does not contain an occlusion map, the comparison is not fair. Therefore, we train a new network of SepConv with an occlusion map.
	\item Ours-\emph{sdc}: To compare our algorithm with kernel and flow combined approaches, we exploit Spatially Displaced Convolution~(SDC)~\cite{reda2018sdc} instead of AdaCoF.
	\item Ours-\emph{ws}: One of the differences between deformable convolution and AdaCoF is that our algorithm does not share the weights over all locations of images. Therefore, we compare it with the weight shared version.
	\item Ours-\emph{woocc}: AdaCoF without occlusion map. The intermediate frame is obtained by simply averaging the outputs from the forward and backward warping.
\end{itemize}
\noindent As shown in Table~\ref{tbl:ablation}, our warping operation outperforms the other ones with lower DoFs. Especially, we can find that the PSNR gap between Ours-\emph{sdc} and Ours is larger than the gap between Ours-\emph{kb} and Ours-\emph{sdc}. It means that breaking the square-shaped kernels to be any shape is more crucial than allowing the kernels to move freely.

\noindent\textbf{Perceptual Loss.} We add perceptual loss $\mathcal{L}_{vgg}$ introduced in Section~\ref{objective} without adversarial loss. We set $\lambda_{vgg}=0.01$. The row of Ours-\emph{vgg} in Table~\ref{tbl:ablation} shows that the PSNR generally decreases and increases only for DAVIS datasets. This implies that the perceptual loss improves the robustness for hard sequences with large and complex motions.

\noindent\textbf{Kernel Size.} We train the network with various kernel sizes $F \in \{1,3,5,7,9,11\}$ which means that $F^2$ offset vectors are used. As shown in Table~\ref{tbl:kernelsize}, the larger kernel size generally leads to better performance and the PSNR saturates as $F$ increases. Especially, the saturation is earlier for the UCF101 dataset because it contains relatively small motion and low-resolution sequences so that there is no room for the performance increase.

\begin{figure}
	\vspace*{-10pt}
	\setlength{\belowcaptionskip}{-24pt}
	\begin{center}
		\subfloat[Ours-$\mathcal{L}_d$]
		{\includegraphics[width=0.485\linewidth]{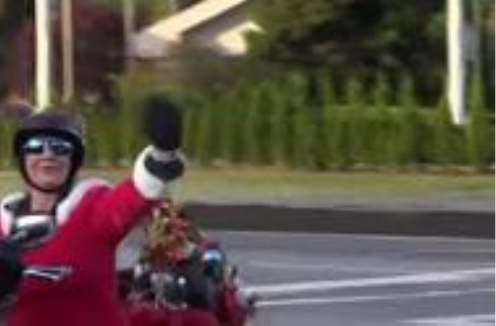}}\,
		\subfloat[Ours-$\mathcal{L}_p$]
		{\includegraphics[width=0.485\linewidth]{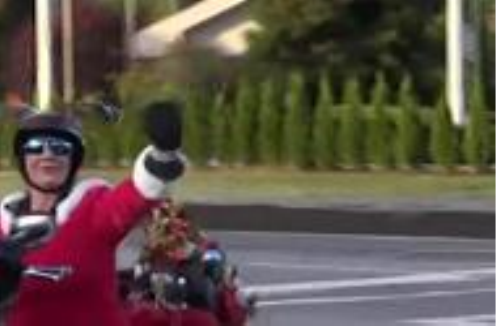}}\	
		\\[-1ex]
		\subfloat[WGAN-GP]
		{\includegraphics[width=0.485\linewidth]{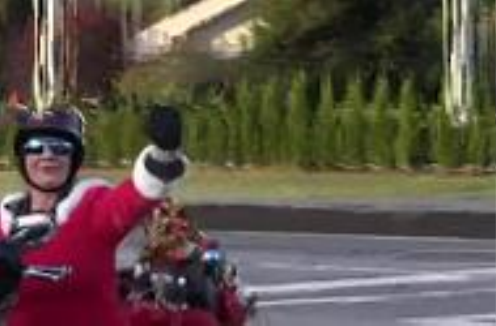}}\,
		\subfloat[TGAN]
		{\includegraphics[width=0.485\linewidth]{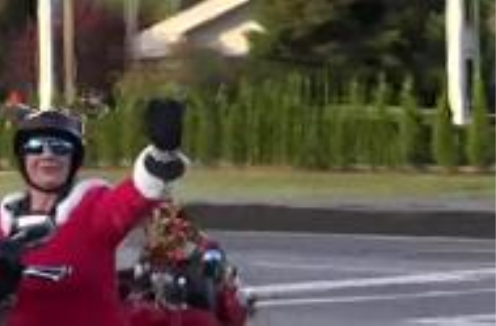}}\,
		\caption{The result of adding adversarial losses.}
		\label{fig:gan}
	\end{center}
\end{figure}

\noindent\textbf{Dilation.} In Section~\ref{network}, we add dilation to the AdaCoF operation to enforce the offset vectors to start from a wider area. We check the effect of dilation by training the network with $F=5$ and $d \in \{0, 1, 2\}$. $d=0$ means that the offset vectors start from the same location. Table~\ref{tbl:dilation} shows that the larger dilation generally leads to better results. As we can see from the \nth{4} - \nth{7} columns of Figure~\ref{fig:offset}, the offset vectors tend to spread more in the case of large motion. Therefore, dilation provides the effect of \emph{better initialization} for them. Figure~\ref{fig:offset} will be covered in more detail in Section~\ref{offsetvis}.

\begin{figure*}
	\setlength{\belowcaptionskip}{-5pt}
	
	\captionsetup[subfigure]{labelformat=empty}
	\subfloat
	{\includegraphics[width=0.195\linewidth]{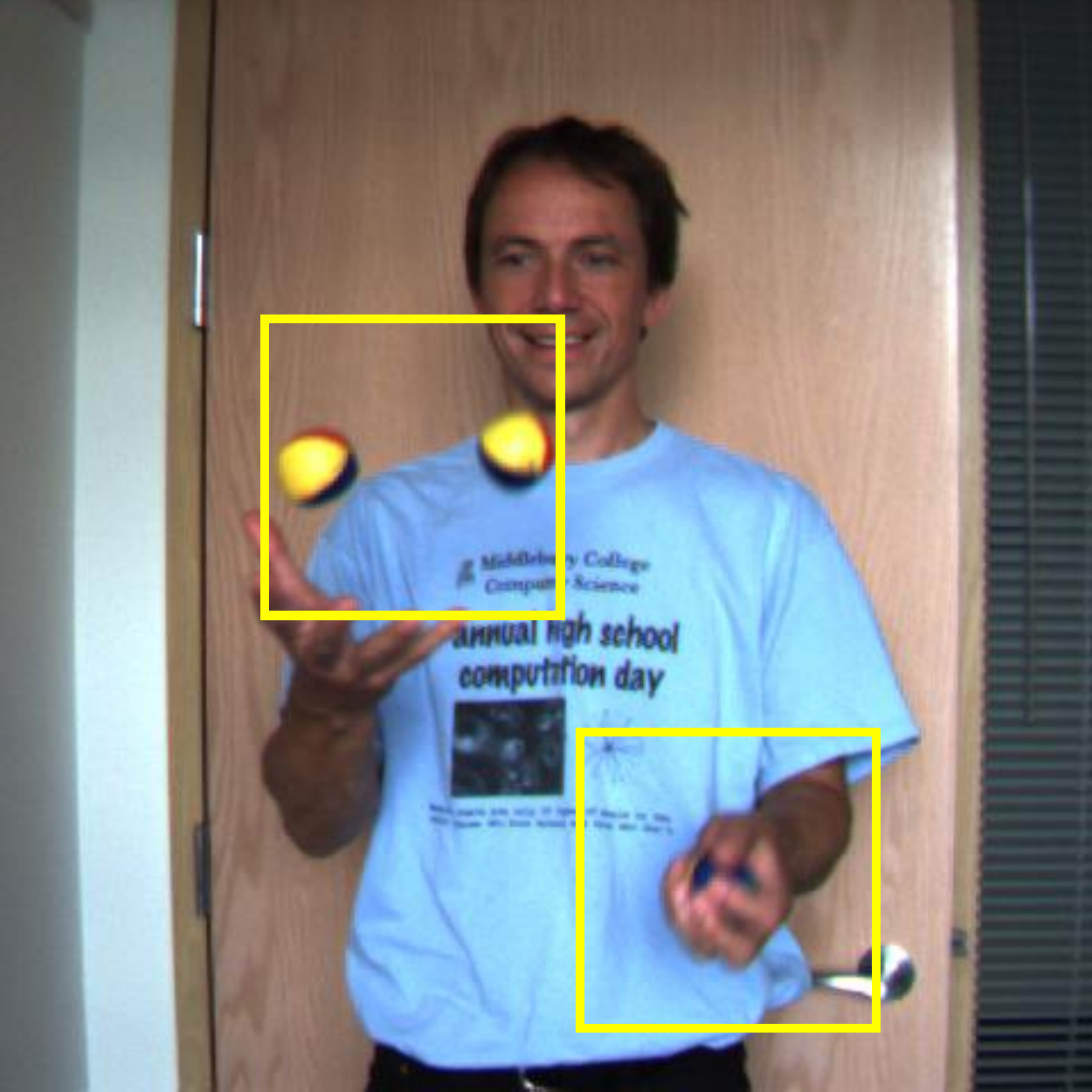}}
	\hspace{-1ex}
	\subfloat{
		\begin{tabular}[b]{c}
			\subfloat
			{\includegraphics[width=0.093\linewidth]{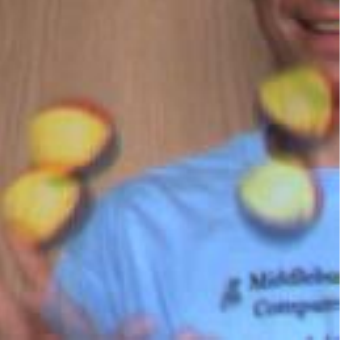}}\
			\subfloat
			{\includegraphics[width=0.093\linewidth]{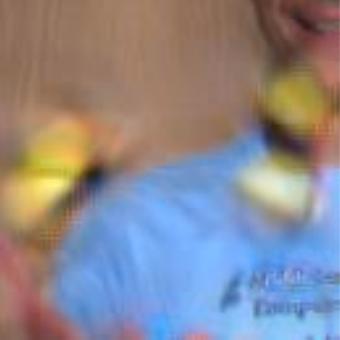}}\
			\subfloat
			{\includegraphics[width=0.093\linewidth]{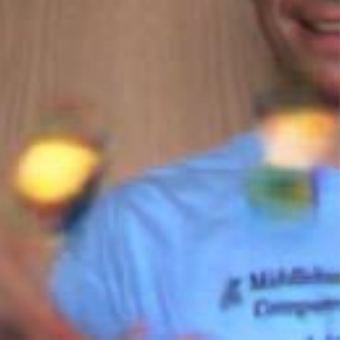}}\
			\subfloat
			{\includegraphics[width=0.093\linewidth]{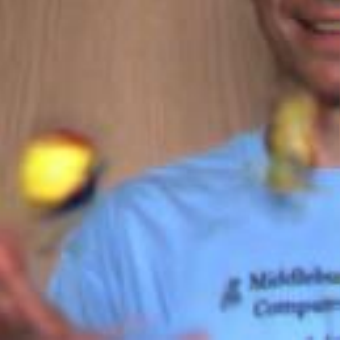}}\
			\subfloat
			{\includegraphics[width=0.093\linewidth]{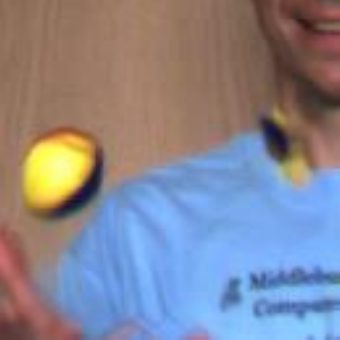}}\
			\subfloat
			{\includegraphics[width=0.093\linewidth]{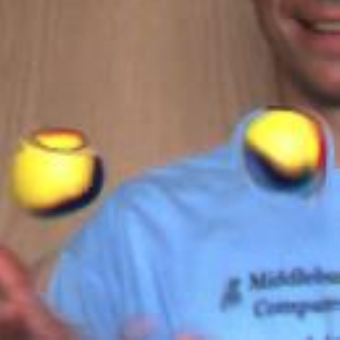}}\
			\subfloat
			{\includegraphics[width=0.093\linewidth]{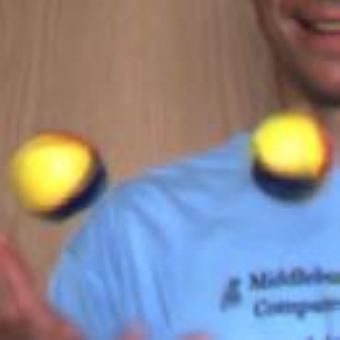}}\
			\subfloat
			{\includegraphics[width=0.093\linewidth]{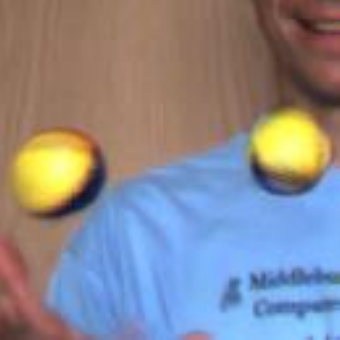}}\
			\\[-2ex]
			\subfloat
			{\includegraphics[width=0.093\linewidth]{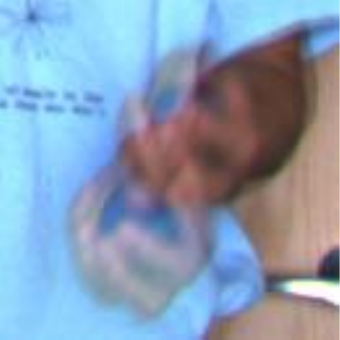}}\
			\subfloat
			{\includegraphics[width=0.093\linewidth]{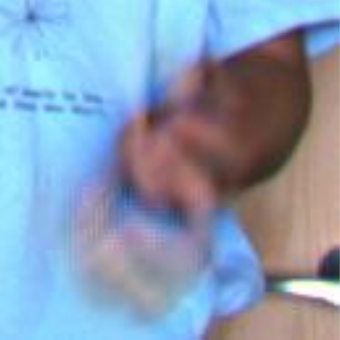}}\
			\subfloat
			{\includegraphics[width=0.093\linewidth]{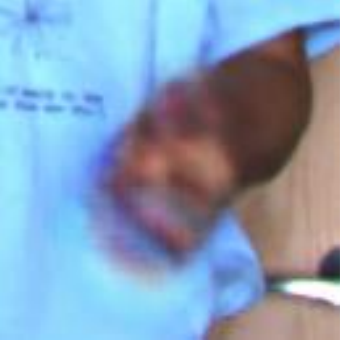}}\
			\subfloat
			{\includegraphics[width=0.093\linewidth]{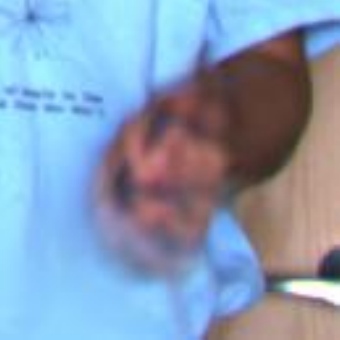}}\
			\subfloat
			{\includegraphics[width=0.093\linewidth]{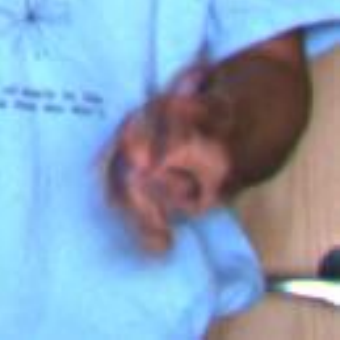}}\
			\subfloat
			{\includegraphics[width=0.093\linewidth]{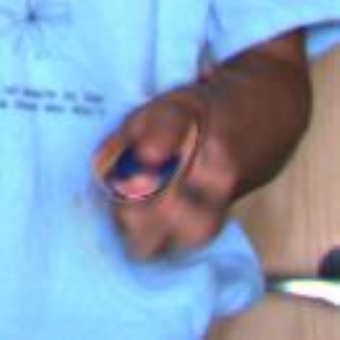}}\
			\subfloat
			{\includegraphics[width=0.093\linewidth]{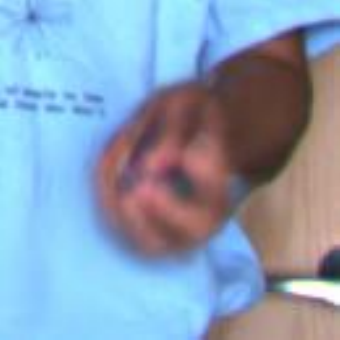}}\
			\subfloat
			{\includegraphics[width=0.093\linewidth]{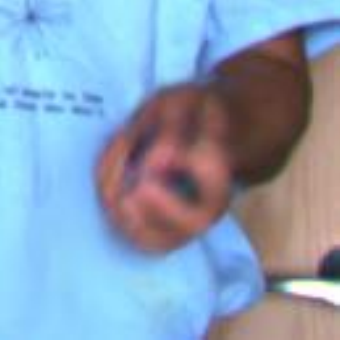}}\
		\end{tabular}
	}
	\\[-4.5ex]
	\subfloat[Ground Truth]
	{\includegraphics[width=0.195\linewidth]{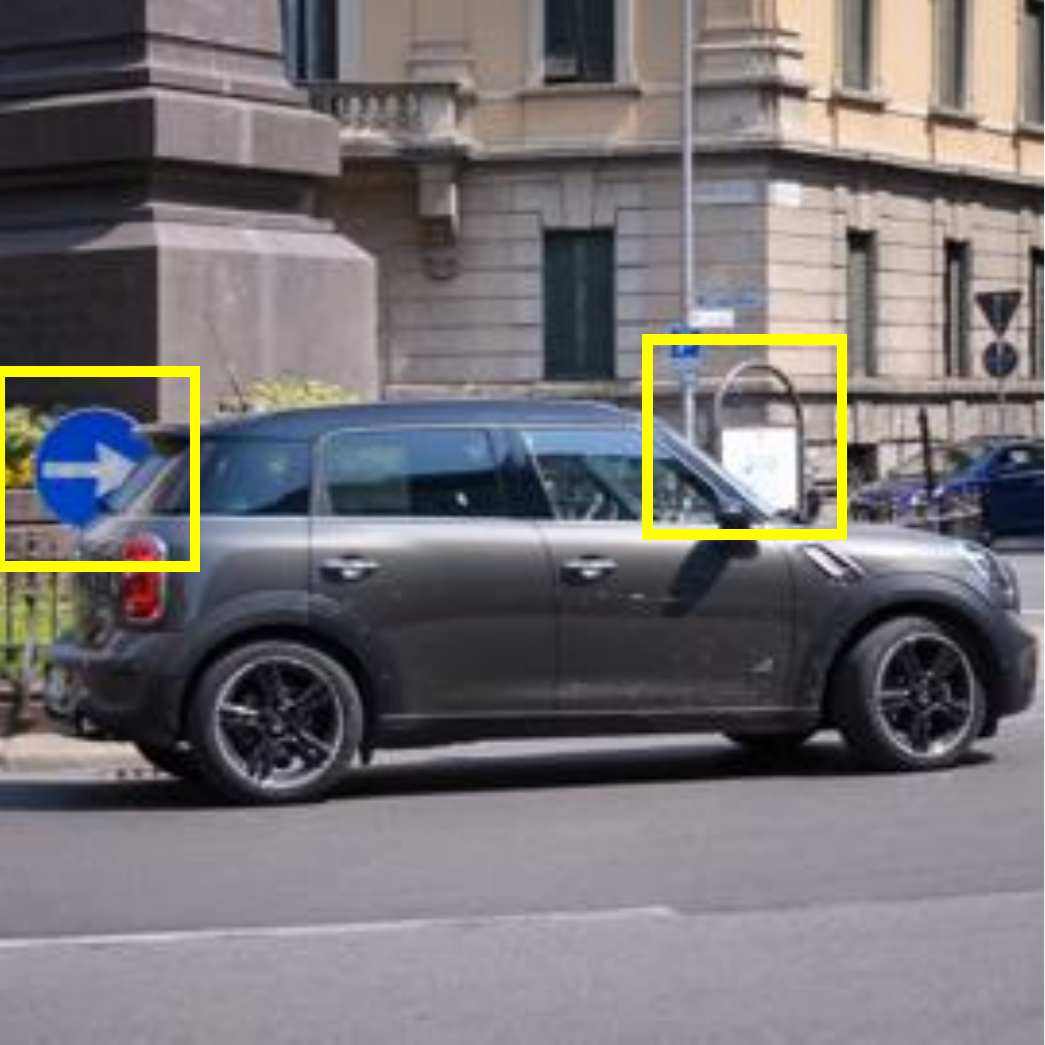}}
	\hspace{-1ex}
	\subfloat{
		\begin{tabular}[b]{c}
			\subfloat
			{\includegraphics[width=0.093\linewidth]{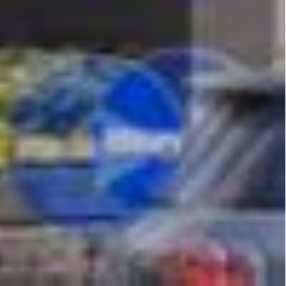}}\
			\subfloat
			{\includegraphics[width=0.093\linewidth]{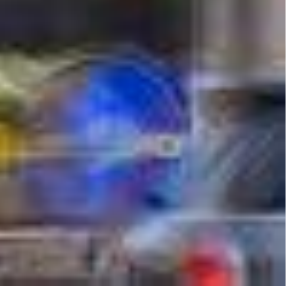}}\
			\subfloat
			{\includegraphics[width=0.093\linewidth]{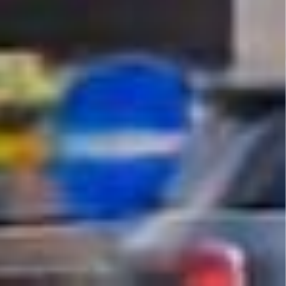}}\
			\subfloat
			{\includegraphics[width=0.093\linewidth]{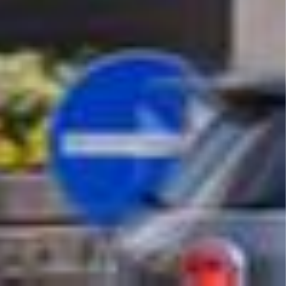}}\
			\subfloat
			{\includegraphics[width=0.093\linewidth]{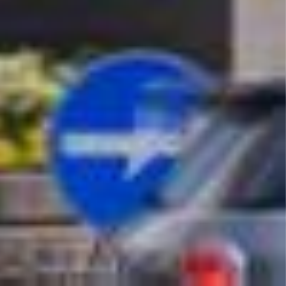}}\
			\subfloat
			{\includegraphics[width=0.093\linewidth]{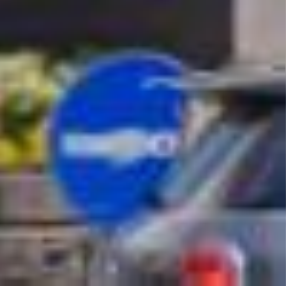}}\
			\subfloat
			{\includegraphics[width=0.093\linewidth]{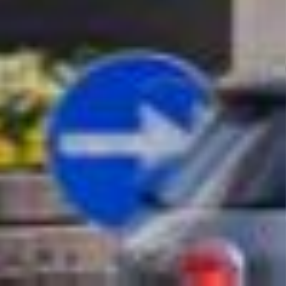}}\
			\subfloat
			{\includegraphics[width=0.093\linewidth]{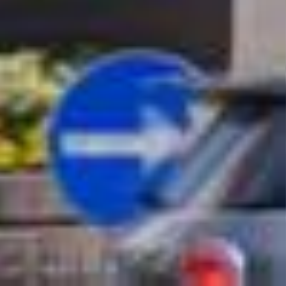}}\
			\\[-2ex]
			\subfloat[Overlap]
			{\includegraphics[width=0.093\linewidth]{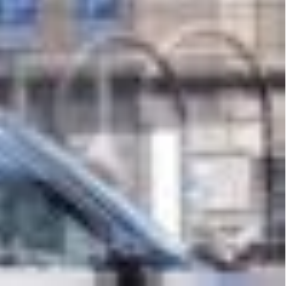}}\
			\subfloat[Phase Based]
			{\includegraphics[width=0.093\linewidth]{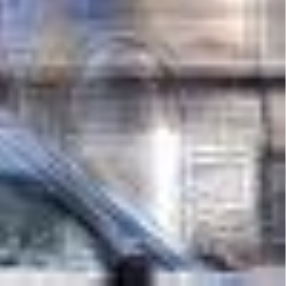}}\
			\subfloat[MIND]
			{\includegraphics[width=0.093\linewidth]{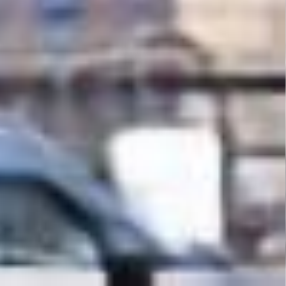}}\
			\subfloat[SepConv]
			{\includegraphics[width=0.093\linewidth]{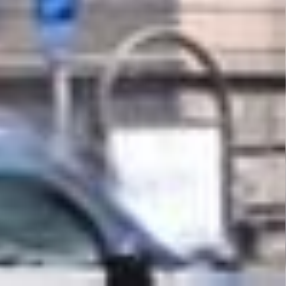}}\
			\subfloat[DVF]
			{\includegraphics[width=0.093\linewidth]{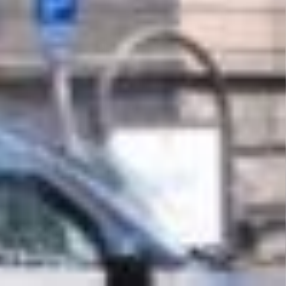}}\
			\subfloat[SuperSlomo]
			{\includegraphics[width=0.093\linewidth]{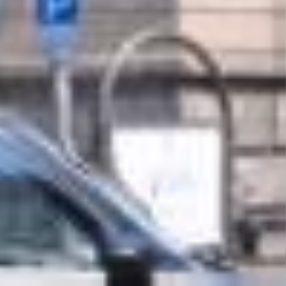}}\
			\subfloat[Ours-$\mathcal{L}_d$]
			{\includegraphics[width=0.093\linewidth]{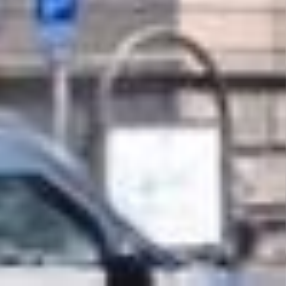}}\
			\subfloat[Ours-$\mathcal{L}_p$]
			{\includegraphics[width=0.093\linewidth]{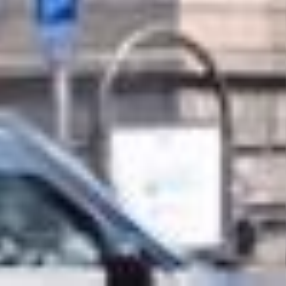}}\
		\end{tabular}
	}
	\caption{Visual comparison of sample sequences with large motions~(\nth{1} - \nth{2} row) and visual comparison of sample sequences with occlusion~(\nth{3} - \nth{4} row). There are occluded areas in the front and back of the car.}
	\label{fig:visual}
	
\end{figure*}

\noindent\textbf{Adversarial Loss.} For the visually more convincing results, we first train the network with $\mathcal{L}_d$ for 50 epochs and fine-tune it for 10 epochs with $\mathcal{L}_p$ which is introduced in Section~\ref{objective}. We set $\lambda_1=0.01, \lambda_{vgg}=1, \lambda_{adv}=0.005$. For the comparison, we train the version of changing $\mathcal{L}_{adv}$ to be WGAN-GP loss~\cite{gulrajani2017improved} and TGAN loss~\cite{saito2017temporal}. Then we visually compare them with the result of the proposed dual-frame adversarial loss~(Ours-$\mathcal{L}_p$). According to Figure~\ref{fig:gan}, fine-tuning the network with adversarial losses increase the sharpness of the results. However, WGAN-GP and TGAN loss cause some artifacts to the output image, while our loss preserves the structures of the frames.

\subsection{Quantitative Evaluation}
\label{evaluation}

We compare our method with simply overlapped results and several competing algorithms including Phase~Based~\cite{meyer2015phase}, MIND~\cite{long2016learning}, SepConv~\cite{sepconv}, DVF~\cite{deepvoxelflow}, and SuperSlomo~\cite{superslomo}. We evaluate two versions of our algorithm. One is the basic version of $F=5, D=1$~(Ours) and the other is the version of $F=11, D=2$~(Ours~+). For a fair comparison, we fix the training environment. We implement the competing algorithms and train them with the train dataset introduced in Section~\ref{implementation} commonly for 50 epochs. We measure PSNR and SSIM of each algorithm for the three test datasets. The results are shown in Table~\ref{tbl:trainfix}. According to the table the kernel-based approach~(SepConv) generally perform better than the flow-based ones~(DVF, SuperSlomo). Finally, our method outperforms the other algorithms for all test datasets by a high margin. We also upload our result to Middlebury Benchmark~\cite{baker2011database} and compare it with the other recent state-of-the-art algorithms. As reported in Table~\ref{tbl:middlebury}, AdaCoF ranks \nth{2} in both IE~(Interpolation Error) and NIE~(Normalized Interpolation Error) among all published methods in Middlebury website. In addition, CyclicGen~\cite{liu2019deep}, which ranks \nth{1} in IE, uses additional edge maps for sharper results and the cycle consistency loss is orthogonally applicable to our method. Also, DAIN~\cite{bao2019depth}, which ranks \nth{1} in NIE, use pre-trained optical flow estimator and depth maps while our method does not require any additional information. Lastly, our approach shows better performance for data with dynamic motions such as Basketball, Dumptruck and Evergreen.

\subsection{Visual Comparison}
\label{viscomp}
Because the video frame interpolation task does not have a fixed answer, the evaluations based on PSNR and SSIM are not perfect by themselves. Therefore we quantitatively evaluate the methods by comparing each result. Especially, we check how our method and other state-of-the-art algorithms handle the two main obstacles which make motions complex in real world videos: large motion and occlusion.

\begin{figure*}
    \vspace*{-10pt}
	\setlength{\belowcaptionskip}{-20pt}
	\begin{center}
		\captionsetup[subfigure]{labelformat=empty}
		\subfloat
		{\includegraphics[width=0.138\linewidth]{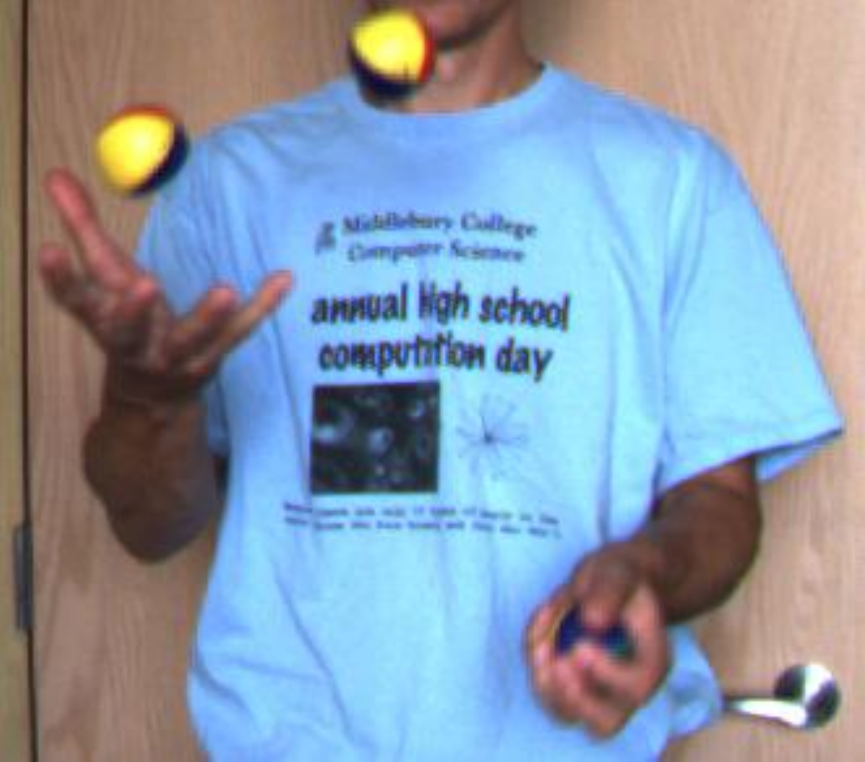}}\
		\hfill
		\subfloat
		{\includegraphics[width=0.138\linewidth]{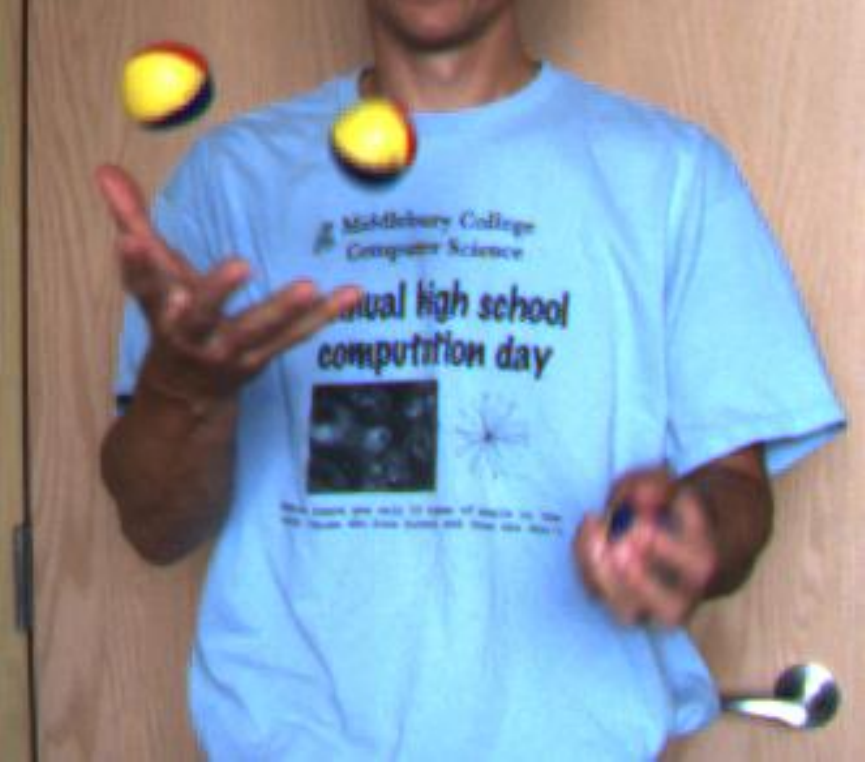}}\
		\hfill
		\subfloat
		{\includegraphics[width=0.138\linewidth]{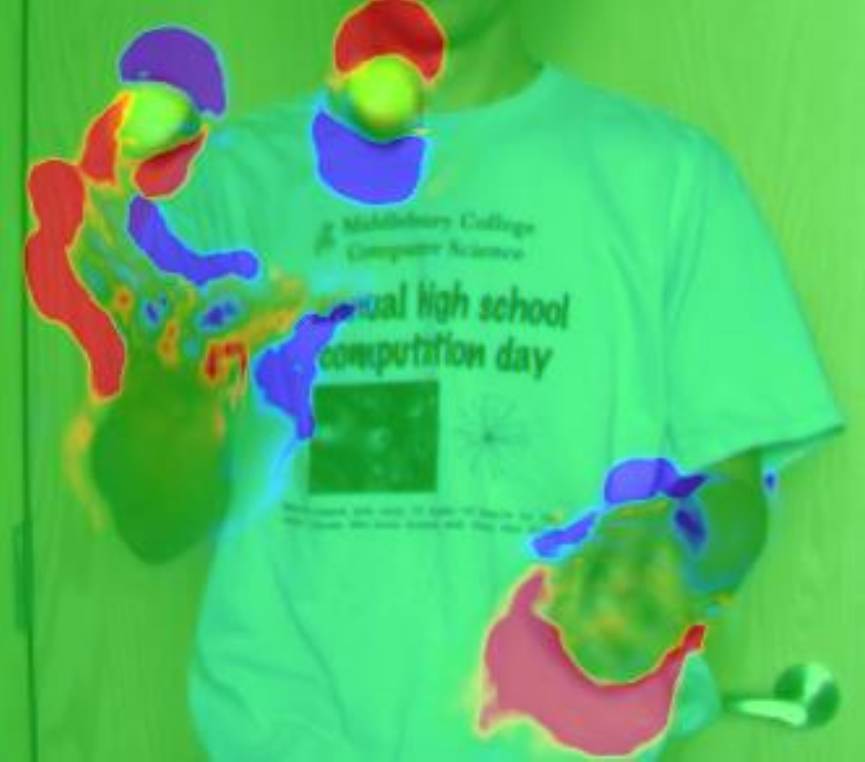}}\
		\hfill
		\subfloat
		{\includegraphics[width=0.138\linewidth]{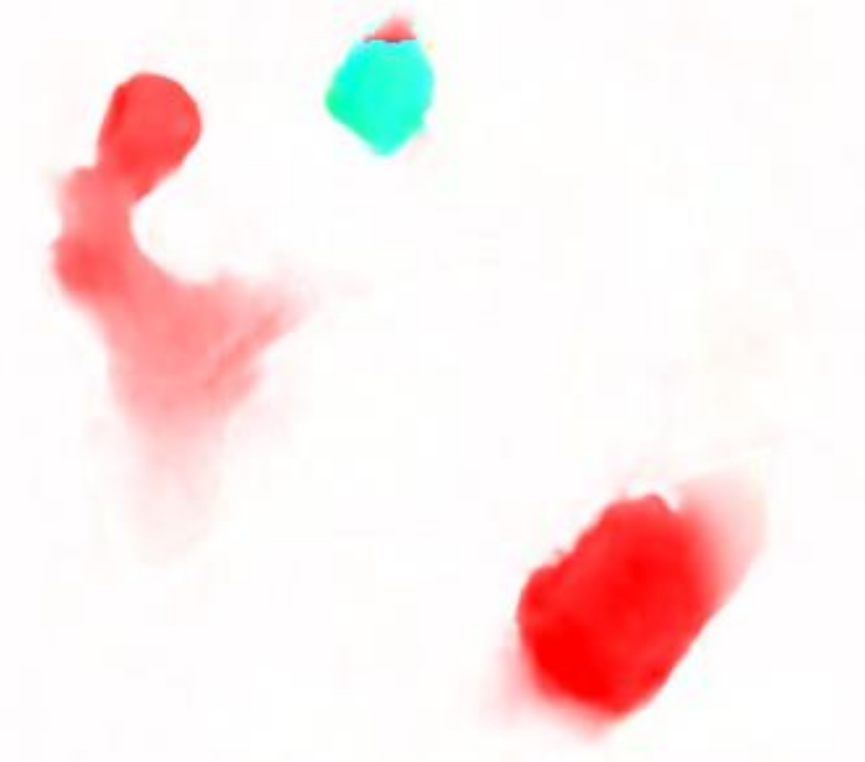}}\
		\hfill
		\subfloat
		{\includegraphics[width=0.138\linewidth]{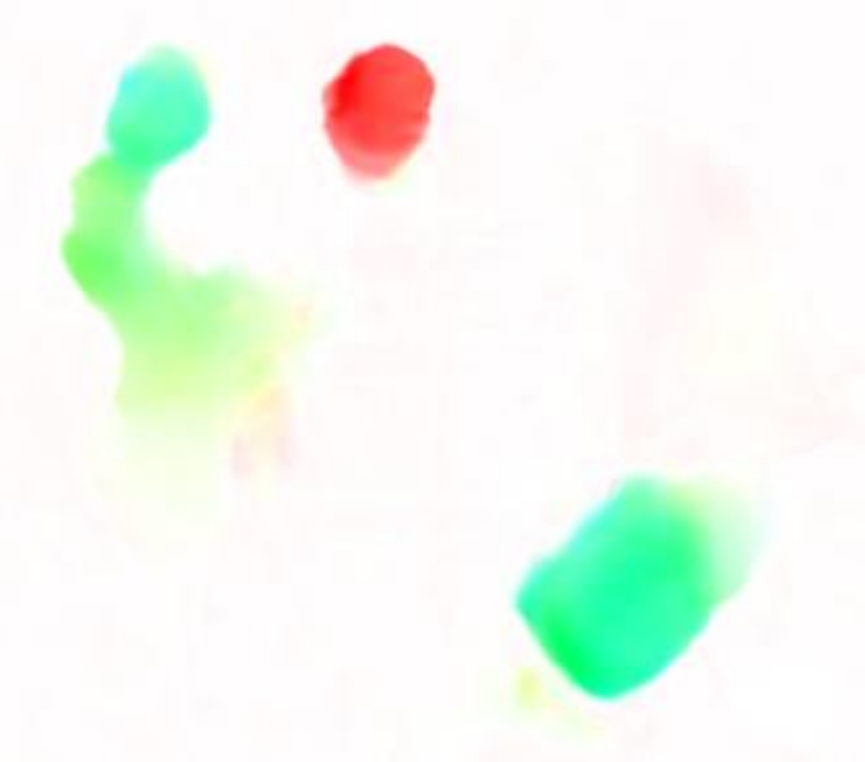}}\
		\hfill
		\subfloat
		{\includegraphics[width=0.138\linewidth]{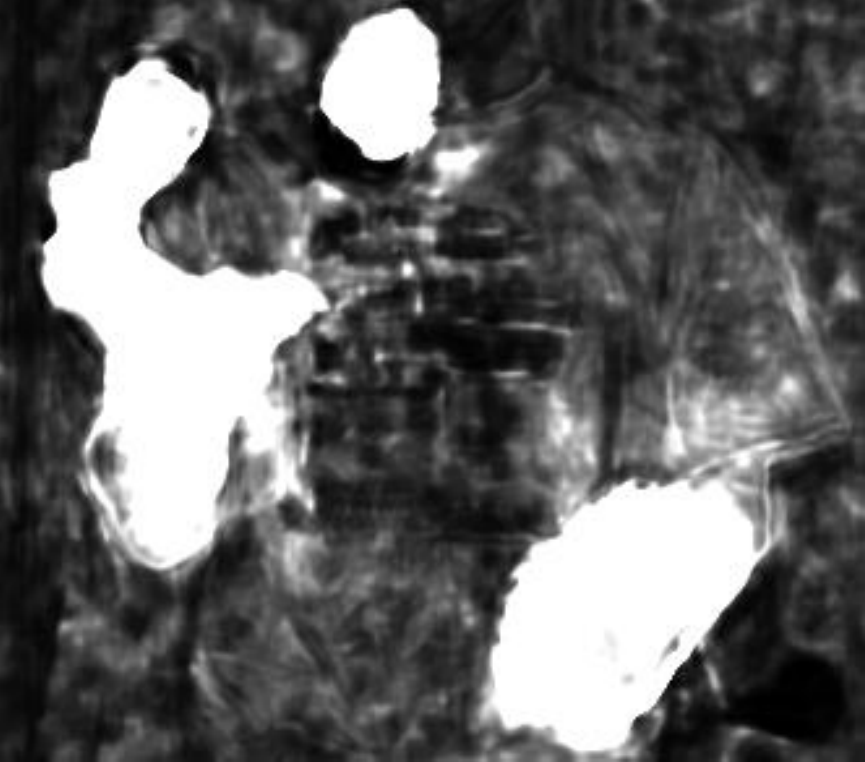}}\
		\hfill
		\subfloat
		{\includegraphics[width=0.138\linewidth]{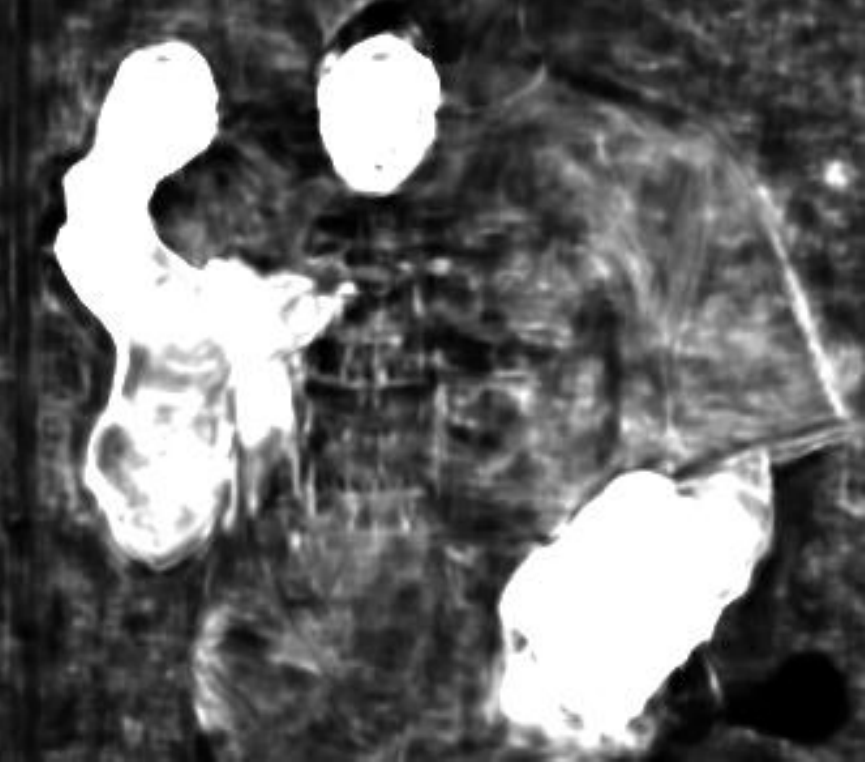}}\
		\\[-2ex]
		\subfloat
		{\includegraphics[width=0.138\linewidth]{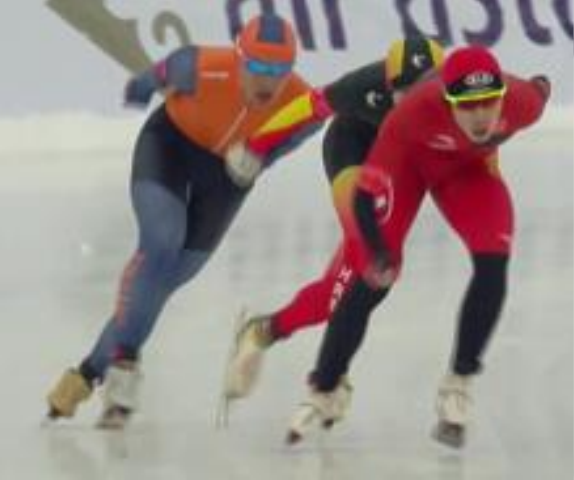}}\
		\hfill
		\subfloat
		{\includegraphics[width=0.138\linewidth]{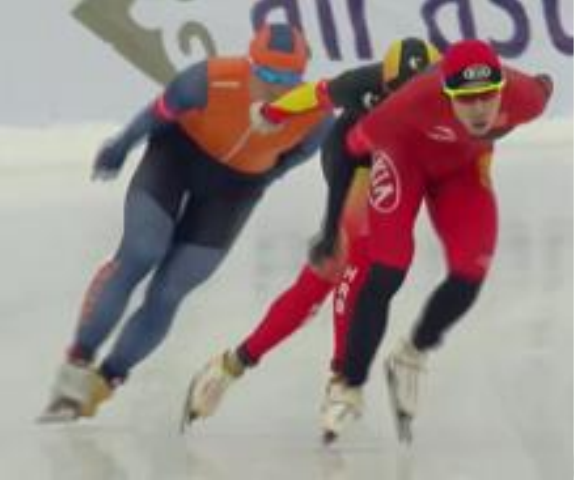}}\
		\hfill
		\subfloat
		{\includegraphics[width=0.138\linewidth]{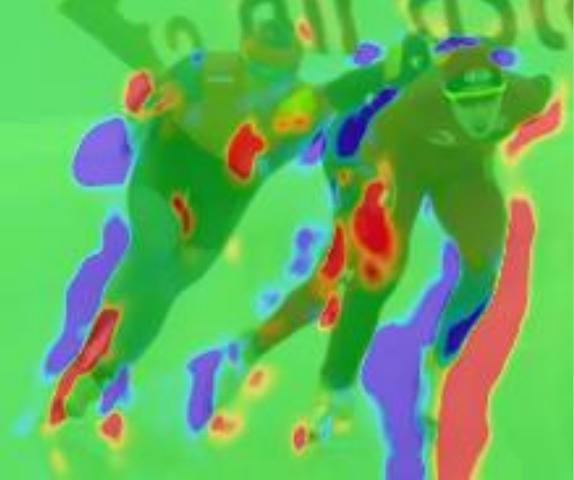}}\
		\hfill
		\subfloat
		{\includegraphics[width=0.138\linewidth]{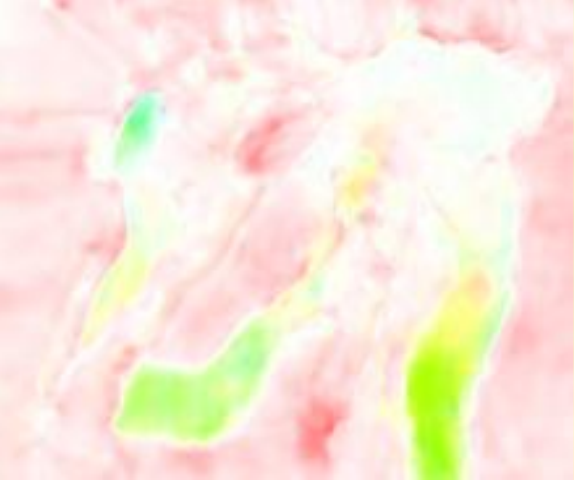}}\
		\hfill
		\subfloat
		{\includegraphics[width=0.138\linewidth]{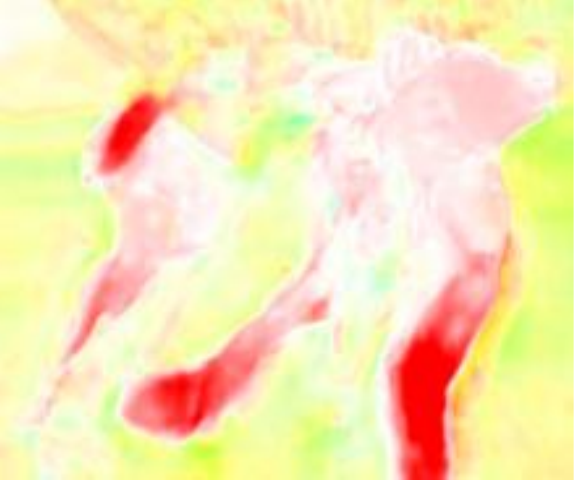}}\
		\hfill
		\subfloat
		{\includegraphics[width=0.138\linewidth]{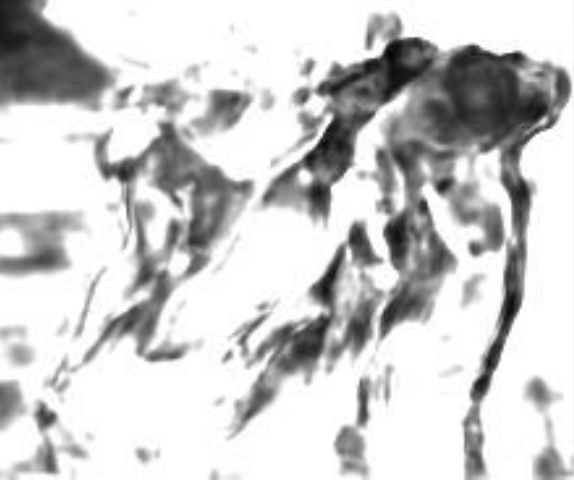}}\
		\hfill
		\subfloat
		{\includegraphics[width=0.138\linewidth]{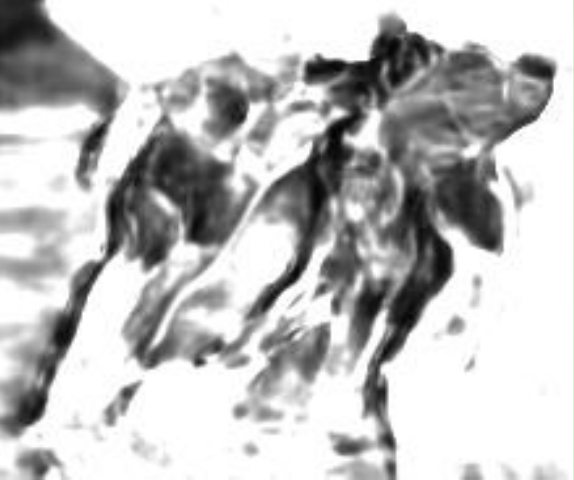}}\
		\\[-2ex]
		\subfloat[Frame1]
		{\includegraphics[width=0.138\linewidth]{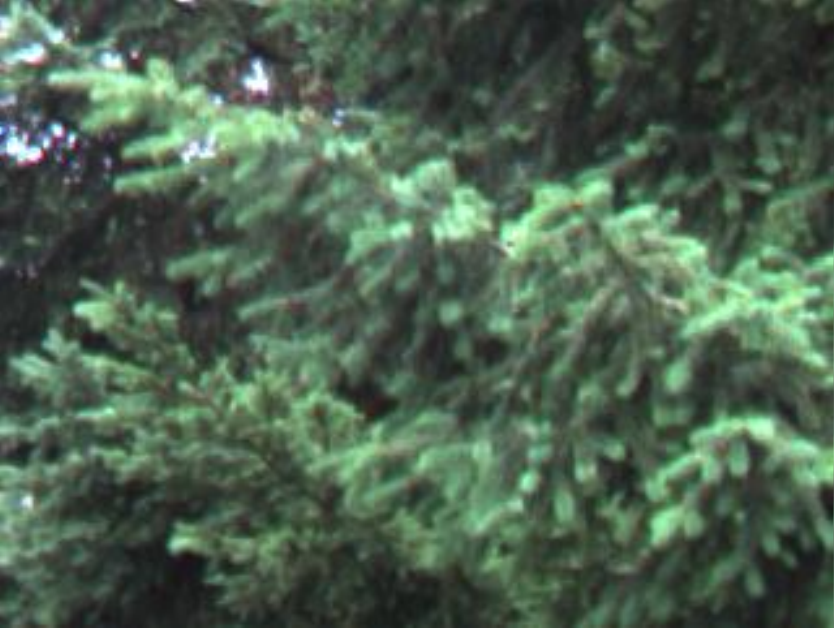}}\
		\hfill
		\subfloat[Frame2]
		{\includegraphics[width=0.138\linewidth]{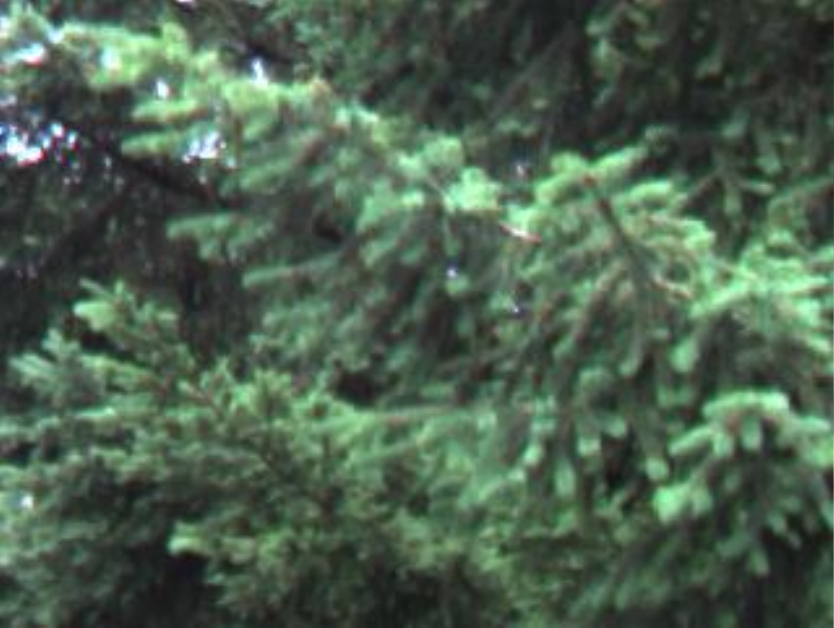}}\
		\hfill
		\subfloat[Occlusion map]
		{\includegraphics[width=0.138\linewidth]{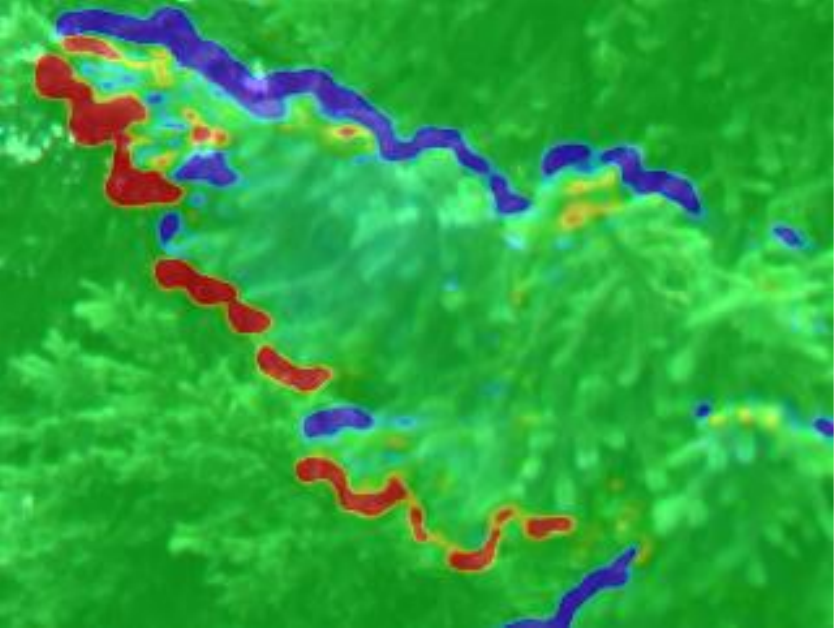}}\
		\hfill
		\subfloat[MeanFlow1]
		{\includegraphics[width=0.138\linewidth]{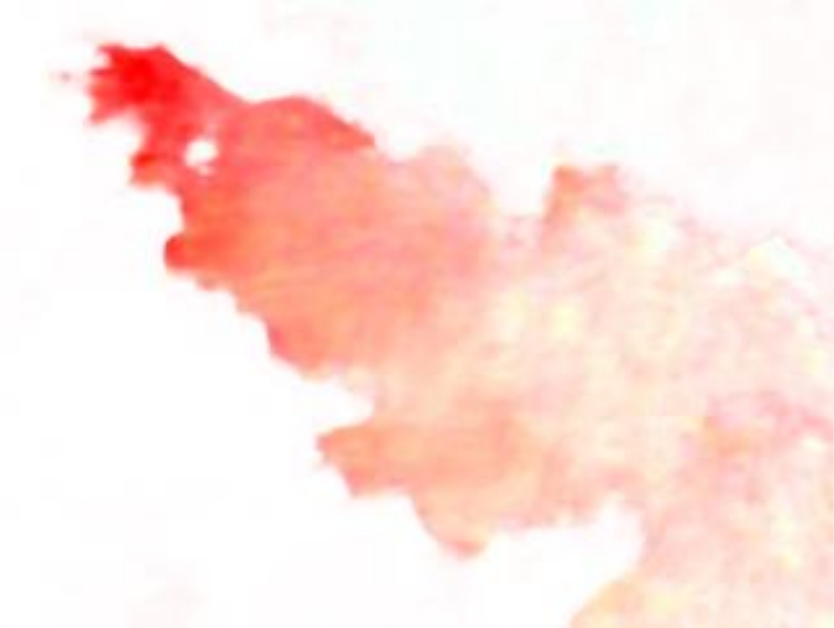}}\
		\hfill
		\subfloat[MeanFlow2]
		{\includegraphics[width=0.138\linewidth]{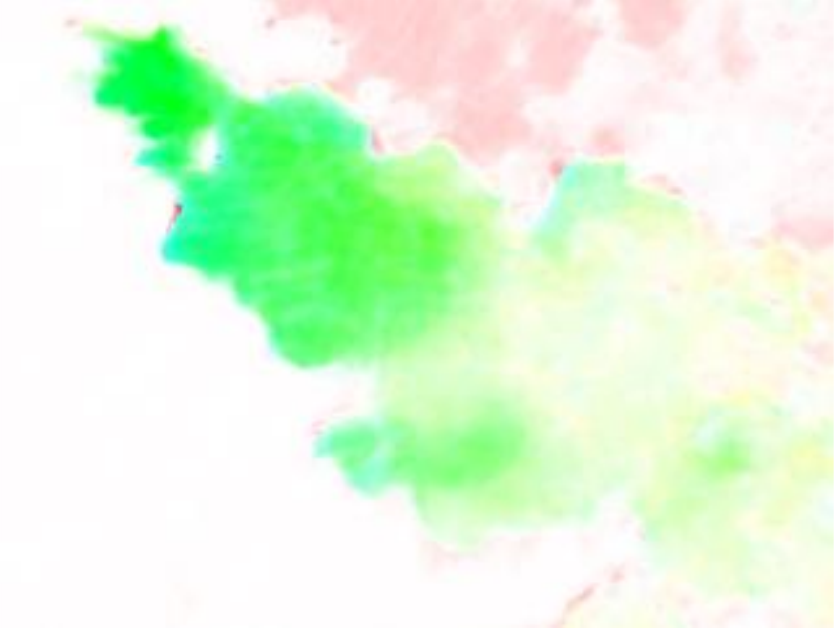}}\
		\hfill
		\subfloat[VarFlow1]
		{\includegraphics[width=0.138\linewidth]{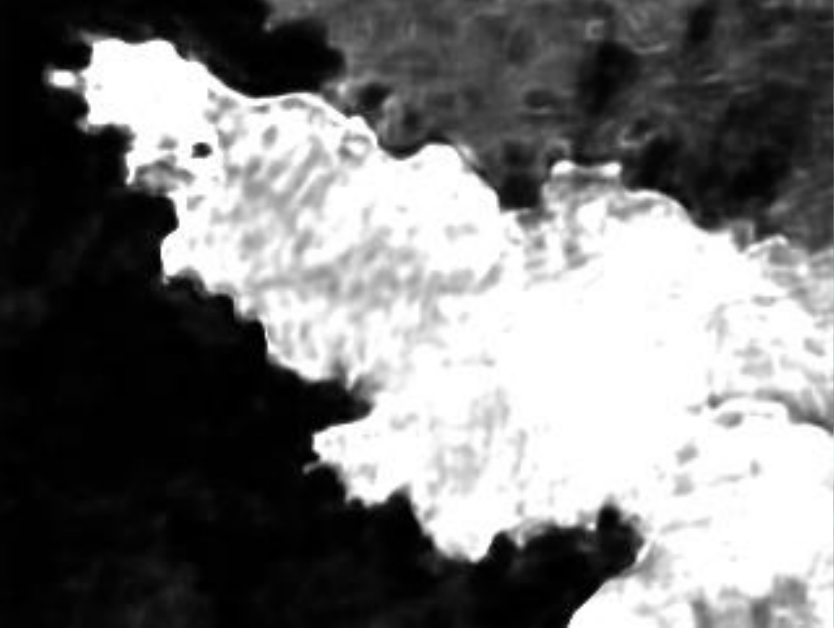}}\
		\hfill
		\subfloat[VarFlow2]
		{\includegraphics[width=0.138\linewidth]{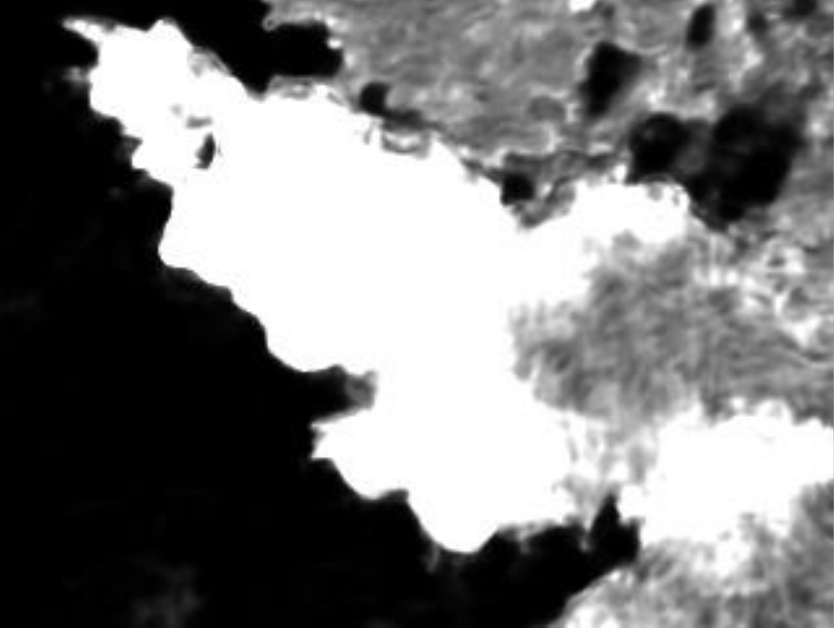}}\
		\vspace*{-5pt}
		\caption{Various visualizations of the network outputs.}
		\label{fig:offset}
	\end{center}
\end{figure*}

\noindent\textbf{Large motion.} When the reference point is located far away, the search area has to be expanded accordingly. Therefore the large motion problem is one of the most challenging obstacles in video frame interpolation research. The first and second rows of Figure~\ref{fig:visual} show the estimated results of various approaches including our method. The results of MIND, SepConv tend to be blurry and DVF, SuperSlomo suffer from some artifacts. Compared to the other competing algorithms, our approach better synthesizes fast moving objects. In addition, the perception-oriented AdaCoF~(Ours-$\mathcal{L}_p$) mitigate the motion blurs of the objects.

\noindent\textbf{Occlusion.} Most of the objects in the intermediate frame appear in both adjacent frames. However, in case of occlusion, the object does not appear in one of the frames. Therefore, the appropriate frame has to be selected for each case, which makes the problem more difficult. In the third and fourth rows of Figure \ref{fig:visual}, a car causes occlusion in its front and back. Comparing the estimated images on occluded areas, our method handles the occlusion problems better than the other approaches.

\subsection{Offset Visualization}
\label{offsetvis}

Our method estimates some parameters from the input images: the kernel weights $W_{k,l}$, the offset vectors $(\alpha_{k,l}, \beta_{k,l})$, and the occlusion map $V$. To check whether the parameters behave as intended, we visualize them in various ways. Further, because the network is trained by self-supervised learning, the visualizations can be obtained without any supervision. Therefore, they can be used for some other tasks in motion estimation research. 

\noindent\textbf{Occlusion map.} The third column of Figure \ref{fig:offset} shows the occlusion map $V$. To handle occlusion, the proper frame has to be selected in each case. For example, the pixels in the red area cannot be found in the second frame. Therefore the network decides to consider only the first frame, not the second one. The blue area can be explained in the same way for the second frame, and the green area means that there is no occlusion.

\noindent\textbf{Mean Flow map.} The fourth and fifth columns of Figure \ref{fig:offset} show the weighted sum of the backward and forward offset vectors for each pixel. We call them Mean Flow $F_m$ and they can be calculated by the following equation.

\begin{equation}
\Delta p_{k,l} = (\alpha_{k,l}, \beta_{k,l})
\end{equation}
\begin{equation}
F_m(i,j) = \sum_{k=0}^{F-1}{\sum_{l=0}^{F-1}{W_{k,l}(i,j)\Delta p_{k,l}}}
\end{equation}

\noindent This means the overall tendency of the offset vectors. Therefore they might behave like a forward/backward optical flow and the figures prove it. This can be used as dense optical flow and can also be obtained from the other flow-based algorithms such as DVF and SuperSlomo.

\noindent\textbf{Variance Flow map.} The sixth and seventh columns of Figure \ref{fig:offset} are the weighted variance of the backward and forward offset vectors. We call them Variance Flow map $F_v$ and they can be calculated by the following equation.

\begin{equation}
F_v(i,j) = \sum_{k=0}^{F-1}{\sum_{l=0}^{F-1}{W_{k,l}(i,j)(F_m(i,j)-\Delta p_{k,l})^2}}
\end{equation}

\noindent The large value for this map means that the offset vectors for the pixel are more spread out so that it can refer to more pixels. According to the figure, more challenging locations such as large motions and occluded areas have larger variance values. Therefore, it can be used as a kind of uncertainty map for some motion estimation tasks. Unlike Mean Flow map, it can only be obtained through our method.

\section{Conclusion}
In this paper, we point out that the DoF of the warping operation to deal with various complex motions is one of the most critical factors in video frame interpolation. Then we propose a new operation called Adaptive Collaboration of Flows~(AdaCoF). This method is the most generalized because all of the previous approaches are special versions of AdaCoF. The parameters needed for the AdaCoF operation are obtained from a fully convolutional network which is end-to-end trainable. Our experiments show that our method outperforms most of the competing algorithms even in several challenging cases such as those with large motion and occlusion. We visualize the network outputs to check whether they behave as intended and that the visualized maps are meaningful, so they can be used for other motion estimation tasks.

\vspace{5mm}

\noindent\textbf{Acknowledgement} This research was supported by Multi-Ministry Collaborative R\&D Program(R\&D program for complex cognitive technology) through the National Research Foundation of Korea(NRF) funded by MSIT, MOTIE, KNPA(NRF-2018M3E3A1057289)

\newpage
\newpage

{\small
	\bibliographystyle{ieee_fullname}
	\bibliography{egbib}

\begin{thebibliography}{10}\itemsep=-1pt

\bibitem{baker2011database}
Simon Baker, Daniel Scharstein, JP Lewis, Stefan Roth, Michael~J Black, and
  Richard Szeliski.
\newblock A database and evaluation methodology for optical flow.
\newblock {\em International Journal of Computer Vision}, 92(1):1--31, 2011.

\bibitem{bao2019depth}
Wenbo Bao, Wei-Sheng Lai, Chao Ma, Xiaoyun Zhang, Zhiyong Gao, and Ming-Hsuan
  Yang.
\newblock Depth-aware video frame interpolation.
\newblock In {\em Proceedings of the IEEE Conference on Computer Vision and
  Pattern Recognition}, pages 3703--3712, 2019.

\bibitem{bao2019memc}
Wenbo Bao, Wei-Sheng Lai, Xiaoyun Zhang, Zhiyong Gao, and Ming-Hsuan Yang.
\newblock Memc-net: Motion estimation and motion compensation driven neural
  network for video interpolation and enhancement.
\newblock {\em IEEE Transactions on Pattern Analysis and Machine Intelligence},
  2019.

\bibitem{barron1994performance}
John~L Barron, David~J Fleet, and Steven~S Beauchemin.
\newblock Performance of optical flow techniques.
\newblock {\em International journal of computer vision}, 12(1):43--77, 1994.

\bibitem{blau2018perception}
Yochai Blau and Tomer Michaeli.
\newblock The perception-distortion tradeoff.
\newblock In {\em Proceedings of the IEEE Conference on Computer Vision and
  Pattern Recognition}, pages 6228--6237, 2018.

\bibitem{chen2016single}
Weifeng Chen, Zhao Fu, Dawei Yang, and Jia Deng.
\newblock Single-image depth perception in the wild.
\newblock In {\em Advances in neural information processing systems}, pages
  730--738, 2016.

\bibitem{chetlur2014cudnn}
Sharan Chetlur, Cliff Woolley, Philippe Vandermersch, Jonathan Cohen, John
  Tran, Bryan Catanzaro, and Evan Shelhamer.
\newblock cudnn: Efficient primitives for deep learning.
\newblock {\em arXiv preprint arXiv:1410.0759}, 2014.

\bibitem{dai2017deformable}
Jifeng Dai, Haozhi Qi, Yuwen Xiong, Yi Li, Guodong Zhang, Han Hu, and Yichen
  Wei.
\newblock Deformable convolutional networks.
\newblock In {\em The IEEE International Conference on Computer Vision (ICCV)},
  Oct 2017.

\bibitem{denton2017unsupervised}
Emily~L Denton et~al.
\newblock Unsupervised learning of disentangled representations from video.
\newblock In {\em Advances in neural information processing systems}, pages
  4414--4423, 2017.

\bibitem{dong2016image}
Chao Dong, Chen~Change Loy, Kaiming He, and Xiaoou Tang.
\newblock Image super-resolution using deep convolutional networks.
\newblock {\em IEEE transactions on pattern analysis and machine intelligence},
  38(2):295--307, 2016.

\bibitem{dosovitskiy2016generating}
Alexey Dosovitskiy and Thomas Brox.
\newblock Generating images with perceptual similarity metrics based on deep
  networks.
\newblock In {\em Advances in neural information processing systems}, pages
  658--666, 2016.

\bibitem{dosovitskiy2015flownet}
Alexey Dosovitskiy, Philipp Fischer, Eddy Ilg, Philip Hausser, Caner Hazirbas,
  Vladimir Golkov, Patrick van~der Smagt, Daniel Cremers, and Thomas Brox.
\newblock Flownet: Learning optical flow with convolutional networks.
\newblock In {\em The IEEE International Conference on Computer Vision (ICCV)},
  December 2015.

\bibitem{ganin2016domain}
Yaroslav Ganin, Evgeniya Ustinova, Hana Ajakan, Pascal Germain, Hugo
  Larochelle, Fran{\c{c}}ois Laviolette, Mario Marchand, and Victor Lempitsky.
\newblock Domain-adversarial training of neural networks.
\newblock {\em The Journal of Machine Learning Research}, 17(1):2096--2030,
  2016.

\bibitem{gatys2016image}
Leon~A. Gatys, Alexander~S. Ecker, and Matthias Bethge.
\newblock Image style transfer using convolutional neural networks.
\newblock In {\em The IEEE Conference on Computer Vision and Pattern
  Recognition (CVPR)}, June 2016.

\bibitem{goodfellow2014generative}
Ian Goodfellow, Jean Pouget-Abadie, Mehdi Mirza, Bing Xu, David Warde-Farley,
  Sherjil Ozair, Aaron Courville, and Yoshua Bengio.
\newblock Generative adversarial nets.
\newblock In {\em Advances in neural information processing systems}, pages
  2672--2680, 2014.

\bibitem{NIPS2015_5951}
Ross Goroshin, Michael~F Mathieu, and Yann LeCun.
\newblock Learning to linearize under uncertainty.
\newblock In C. Cortes, N.~D. Lawrence, D.~D. Lee, M. Sugiyama, and R. Garnett,
  editors, {\em Advances in Neural Information Processing Systems 28}, pages
  1234--1242. Curran Associates, Inc., 2015.

\bibitem{gulrajani2017improved}
Ishaan Gulrajani, Faruk Ahmed, Martin Arjovsky, Vincent Dumoulin, and Aaron~C
  Courville.
\newblock Improved training of wasserstein gans.
\newblock In {\em Advances in neural information processing systems}, pages
  5767--5777, 2017.

\bibitem{he2016deep}
Kaiming He, Xiangyu Zhang, Shaoqing Ren, and Jian Sun.
\newblock Deep residual learning for image recognition.
\newblock In {\em The IEEE Conference on Computer Vision and Pattern
  Recognition (CVPR)}, June 2016.

\bibitem{ilg2017flownet}
Eddy Ilg, Nikolaus Mayer, Tonmoy Saikia, Margret Keuper, Alexey Dosovitskiy,
  and Thomas Brox.
\newblock Flownet 2.0: Evolution of optical flow estimation with deep networks.
\newblock In {\em IEEE conference on computer vision and pattern recognition
  (CVPR)}, volume~2, page~6, 2017.

\bibitem{superslomo}
Huaizu Jiang, Deqing Sun, Varun Jampani, Ming-Hsuan Yang, Erik Learned-Miller,
  and Jan Kautz.
\newblock Super slomo: High quality estimation of multiple intermediate frames
  for video interpolation.
\newblock In {\em The IEEE Conference on Computer Vision and Pattern
  Recognition (CVPR)}, June 2018.

\bibitem{johnson2016perceptual}
Justin Johnson, Alexandre Alahi, and Li Fei-Fei.
\newblock Perceptual losses for real-time style transfer and super-resolution.
\newblock In {\em European Conference on Computer Vision}, pages 694--711.
  Springer, 2016.

\bibitem{kingma2014adam}
Diederik~P Kingma and Jimmy Ba.
\newblock Adam: A method for stochastic optimization.
\newblock {\em arXiv preprint arXiv:1412.6980}, 2014.

\bibitem{krizhevsky2012imagenet}
Alex Krizhevsky, Ilya Sutskever, and Geoffrey~E Hinton.
\newblock Imagenet classification with deep convolutional neural networks.
\newblock In {\em Advances in neural information processing systems}, pages
  1097--1105, 2012.

\bibitem{ledig2017photo}
Christian Ledig, Lucas Theis, Ferenc Husz{\'a}r, Jose Caballero, Andrew
  Cunningham, Alejandro Acosta, Andrew Aitken, Alykhan Tejani, Johannes Totz,
  Zehan Wang, et~al.
\newblock Photo-realistic single image super-resolution using a generative
  adversarial network.
\newblock In {\em Proceedings of the IEEE conference on computer vision and
  pattern recognition}, pages 4681--4690, 2017.

\bibitem{liu2012multiple}
Hongbin Liu, Ruiqin Xiong, Debin Zhao, Siwei Ma, and Wen Gao.
\newblock Multiple hypotheses bayesian frame rate up-conversion by adaptive
  fusion of motion-compensated interpolations.
\newblock {\em IEEE transactions on circuits and systems for video technology},
  22(8):1188--1198, 2012.

\bibitem{liu2019deep}
Yu-Lun Liu, Yi-Tung Liao, Yen-Yu Lin, and Yung-Yu Chuang.
\newblock Deep video frame interpolation using cyclic frame generation.
\newblock In {\em AAAI Conference on Artificial Intelligence}, 2019.

\bibitem{deepvoxelflow}
Ziwei Liu, Raymond~A. Yeh, Xiaoou Tang, Yiming Liu, and Aseem Agarwala.
\newblock Video frame synthesis using deep voxel flow.
\newblock In {\em The IEEE International Conference on Computer Vision (ICCV)},
  Oct 2017.

\bibitem{long2016learning}
Gucan Long, Laurent Kneip, Jose~M Alvarez, Hongdong Li, Xiaohu Zhang, and
  Qifeng Yu.
\newblock Learning image matching by simply watching video.
\newblock In {\em European Conference on Computer Vision}, pages 434--450.
  Springer, 2016.

\bibitem{mahajan2009moving}
Dhruv Mahajan, Fu-Chung Huang, Wojciech Matusik, Ravi Ramamoorthi, and Peter
  Belhumeur.
\newblock Moving gradients: a path-based method for plausible image
  interpolation.
\newblock In {\em ACM Transactions on Graphics (TOG)}, volume~28, page~42. ACM,
  2009.

\bibitem{mathieu2015deep}
Michael Mathieu, Camille Couprie, and Yann LeCun.
\newblock Deep multi-scale video prediction beyond mean square error.
\newblock In {\em International Conference on Learning Representations (ICLR)},
  2016.

\bibitem{meyer2018phasenet}
Simone Meyer, Abdelaziz Djelouah, Brian McWilliams, Alexander Sorkine-Hornung,
  Markus Gross, and Christopher Schroers.
\newblock Phasenet for video frame interpolation.
\newblock In {\em The IEEE Conference on Computer Vision and Pattern
  Recognition (CVPR)}, June 2018.

\bibitem{meyer2015phase}
Simone Meyer, Oliver Wang, Henning Zimmer, Max Grosse, and Alexander
  Sorkine-Hornung.
\newblock Phase-based frame interpolation for video.
\newblock In {\em The IEEE Conference on Computer Vision and Pattern
  Recognition (CVPR)}, June 2015.

\bibitem{Niklaus_2018_CVPR}
Simon Niklaus and Feng Liu.
\newblock Context-aware synthesis for video frame interpolation.
\newblock In {\em The IEEE Conference on Computer Vision and Pattern
  Recognition (CVPR)}, June 2018.

\bibitem{adaconv}
Simon Niklaus, Long Mai, and Feng Liu.
\newblock Video frame interpolation via adaptive convolution.
\newblock In {\em The IEEE Conference on Computer Vision and Pattern
  Recognition (CVPR)}, July 2017.

\bibitem{sepconv}
Simon Niklaus, Long Mai, and Feng Liu.
\newblock Video frame interpolation via adaptive separable convolution.
\newblock In {\em The IEEE International Conference on Computer Vision (ICCV)},
  Oct 2017.

\bibitem{paszke2017automatic}
Adam Paszke, Sam Gross, Soumith Chintala, Gregory Chanan, Edward Yang, Zachary
  DeVito, Zeming Lin, Alban Desmaison, Luca Antiga, and Adam Lerer.
\newblock Automatic differentiation in pytorch.
\newblock In {\em NIPS 2017 Autodiff Workshop: The Future of Gradient-based
  Machine Learning Software and Techniques}, 2017.

\bibitem{Perazzi2016}
F. Perazzi, J. Pont-Tuset, B. McWilliams, L. {Van Gool}, M. Gross, and A.
  Sorkine-Hornung.
\newblock A benchmark dataset and evaluation methodology for video object
  segmentation.
\newblock In {\em Computer Vision and Pattern Recognition}, 2016.

\bibitem{reda2018sdc}
Fitsum~A Reda, Guilin Liu, Kevin~J Shih, Robert Kirby, Jon Barker, David
  Tarjan, Andrew Tao, and Bryan Catanzaro.
\newblock Sdc-net: Video prediction using spatially-displaced convolution.
\newblock In {\em Proceedings of the European Conference on Computer Vision
  (ECCV)}, pages 718--733, 2018.

\bibitem{10.1007/978-3-319-24574-4_28}
Olaf Ronneberger, Philipp Fischer, and Thomas Brox.
\newblock U-net: Convolutional networks for biomedical image segmentation.
\newblock In Nassir Navab, Joachim Hornegger, William~M. Wells, and
  Alejandro~F. Frangi, editors, {\em Medical Image Computing and
  Computer-Assisted Intervention -- MICCAI 2015}, pages 234--241, Cham, 2015.
  Springer International Publishing.

\bibitem{saito2017temporal}
Masaki Saito, Eiichi Matsumoto, and Shunta Saito.
\newblock Temporal generative adversarial nets with singular value clipping.
\newblock In {\em Proceedings of the IEEE International Conference on Computer
  Vision}, pages 2830--2839, 2017.

\bibitem{simonyan2014very}
Karen Simonyan and Andrew Zisserman.
\newblock Very deep convolutional networks for large-scale image recognition.
\newblock {\em arXiv preprint arXiv:1409.1556}, 2014.

\bibitem{soomro2012ucf101}
Khurram Soomro, Amir~Roshan Zamir, and Mubarak Shah.
\newblock Ucf101: A dataset of 101 human actions classes from videos in the
  wild.
\newblock {\em arXiv preprint arXiv:1212.0402}, 2012.

\bibitem{srivastava2015unsupervised}
Nitish Srivastava, Elman Mansimov, and Ruslan Salakhudinov.
\newblock Unsupervised learning of video representations using lstms.
\newblock In {\em International conference on machine learning}, pages
  843--852, 2015.

\bibitem{sun2018pwc}
Deqing Sun, Xiaodong Yang, Ming-Yu Liu, and Jan Kautz.
\newblock Pwc-net: Cnns for optical flow using pyramid, warping, and cost
  volume.
\newblock In {\em The IEEE Conference on Computer Vision and Pattern
  Recognition (CVPR)}, June 2018.

\bibitem{wang2004image}
Zhou Wang, Alan~C Bovik, Hamid~R Sheikh, Eero~P Simoncelli, et~al.
\newblock Image quality assessment: from error visibility to structural
  similarity.
\newblock {\em IEEE transactions on image processing}, 13(4):600--612, 2004.

\bibitem{weinzaepfel2013deepflow}
Philippe Weinzaepfel, Jerome Revaud, Zaid Harchaoui, and Cordelia Schmid.
\newblock Deepflow: Large displacement optical flow with deep matching.
\newblock In {\em The IEEE International Conference on Computer Vision (ICCV)},
  December 2013.

\bibitem{werlberger2011optical}
Manuel Werlberger, Thomas Pock, Markus Unger, and Horst Bischof.
\newblock Optical flow guided tv-l 1 video interpolation and restoration.
\newblock In {\em International Workshop on Energy Minimization Methods in
  Computer Vision and Pattern Recognition}, pages 273--286. Springer, 2011.

\bibitem{xu2012motion}
Li Xu, Jiaya Jia, and Yasuyuki Matsushita.
\newblock Motion detail preserving optical flow estimation.
\newblock {\em IEEE Transactions on Pattern Analysis and Machine Intelligence},
  34(9):1744--1757, 2012.

\bibitem{xue2019video}
Tianfan Xue, Baian Chen, Jiajun Wu, Donglai Wei, and William~T Freeman.
\newblock Video enhancement with task-oriented flow.
\newblock {\em International Journal of Computer Vision}, 127(8):1106--1125,
  2019.

\bibitem{yu2013multi}
Zhefei Yu, Houqiang Li, Zhangyang Wang, Zeng Hu, and Chang~Wen Chen.
\newblock Multi-level video frame interpolation: Exploiting the interaction
  among different levels.
\newblock {\em IEEE Transactions on Circuits and Systems for Video Technology},
  23(7):1235--1248, 2013.

\bibitem{zhu2016generative}
Jun-Yan Zhu, Philipp Kr{\"a}henb{\"u}hl, Eli Shechtman, and Alexei~A Efros.
\newblock Generative visual manipulation on the natural image manifold.
\newblock In {\em European Conference on Computer Vision (ECCV)}, pages
  597--613. Springer, 2016.

\end{thebibliography}
}

\end{document}